%% file: main.tex
\documentclass[letterpaper]{article} 
\usepackage{aaai21}  
\usepackage{times}  
\usepackage{helvet} 
\usepackage{courier}  
\usepackage[hyphens]{url}  
\usepackage{graphicx} 
\usepackage{natbib}  
\usepackage{caption} 
\usepackage{subcaption}
\usepackage{multirow}

\usepackage[draft=false]{defs-common}
\usepackage{defs-custom}
\begin{document}

\title{Procedural Content Generation using Behavior Trees (PCGBT)}

\author{
Anurag Sarkar and Seth Cooper\\
}

\affiliations{
Northeastern University\\
sarkar.an@northeastern.edu, se.cooper@northeastern.edu
}

\maketitle

\begin{abstract}
Behavior trees (BTs) are a popular method for modeling NPC and enemy AI behavior and have been widely used in commercial games. In this work, rather than use BTs to model game \textit{playing} agents, we use them for modeling game \textit{design} agents, defining behaviors as content generation tasks rather than in-game actions. Similar to how traditional BTs enable modeling behaviors in a modular and dynamic manner, BTs for PCG enable simple subtrees for generating parts of levels to be combined modularly to form complex trees for generating whole levels as well as generators that can dynamically vary the generated content. We refer to this approach as Procedural Content Generation using Behavior Trees, or PCGBT, and demonstrate it by using BTs to model generators for \textit{Super Mario Bros.}, \textit{Mega Man} and \textit{Metroid} levels as well as dungeon layouts and discuss several ways in which this paradigm could be applied and extended in the future.
\end{abstract}

\input{figuretable}
\input{body}


\bibliography{refs-custom}

\end{document}

%% file: figuretable.tex

\newcolumntype{"}{@{\hskip\tabcolsep\vrule width 1pt\hskip\tabcolsep}}

\newcommand{\XFIGUREsmboneone}{
\begin{figure*}[t]
\centering
\begin{tabular}{c}
\includegraphics[width=1.64\columnwidth]{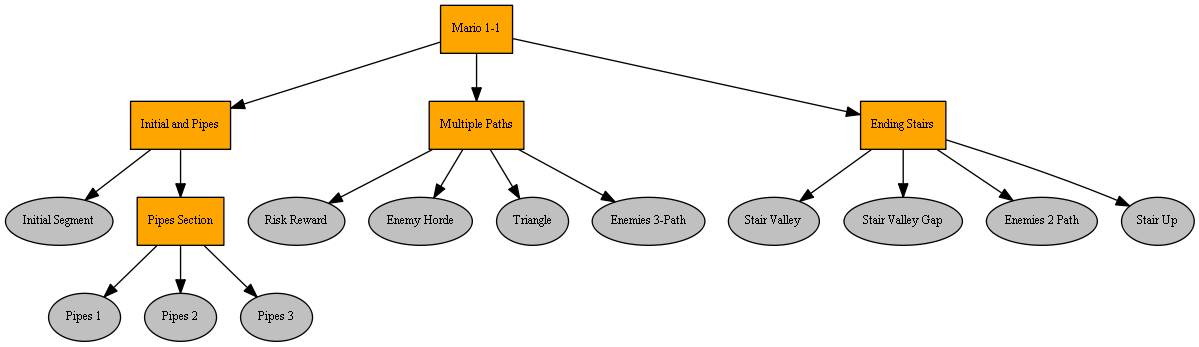}
\\
\raisebox{3pt}{\rotatebox{90}{\scriptsize{verbatim}}}
\includegraphics[width=1.5\columnwidth]{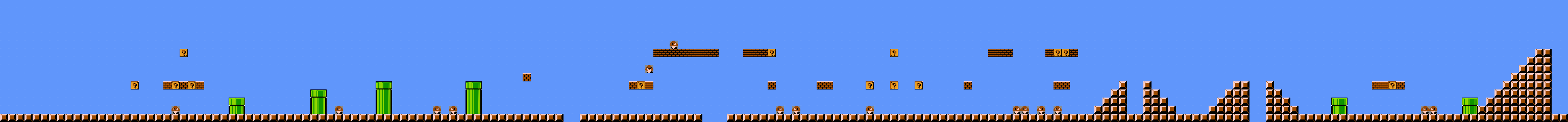}
\\
\raisebox{2pt}{\rotatebox{90}{\scriptsize{sampled}}}
\includegraphics[width=1.5\columnwidth]{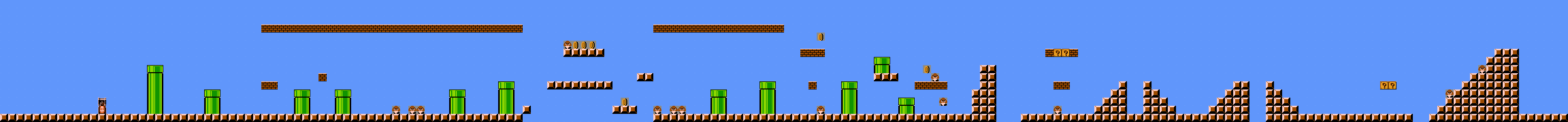}
\end{tabular}
\caption{\label{XFIGUREsmboneone} SMB 1-1 PCG-BT along with a recreation of the original level (verbatim) and a variant (sampled).}
\end{figure*}
}

\newcommand{\XFIGUREsmblevelbtonecol}{
\begin{figure}[t]
\centering
\begin{tabular}{c}
{\includegraphics[width=1.0\columnwidth]{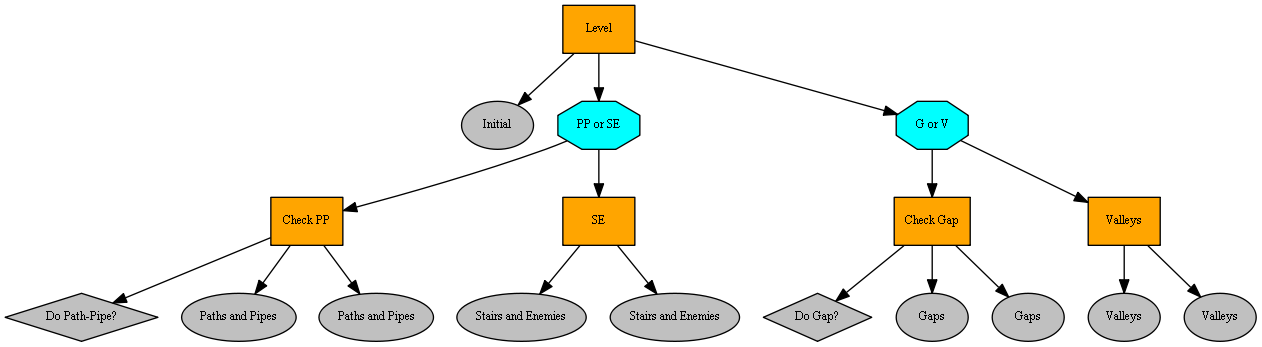}} \\ \includegraphics[width=0.75\columnwidth]{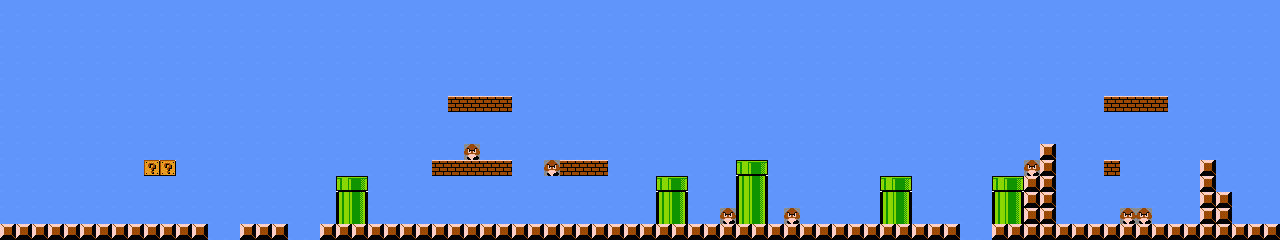} \\ \includegraphics[width=0.75\columnwidth]{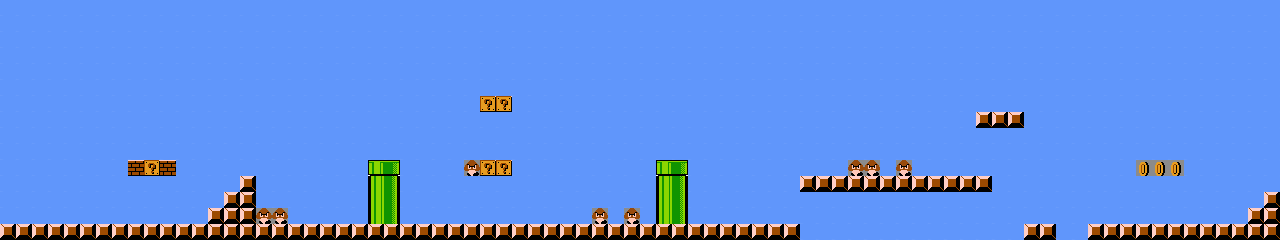}
\end{tabular}
\caption{\label{XFIGUREsmblevelbtonecol} SMB PCG-BT and example generated levels.}
\end{figure}
}

\newcommand{\XFIGUREsmblevelbt}{
\begin{figure*}[t]
\centering
\begin{tabular}{cc}
\multicolumn{2}{c}{{\includegraphics[width=2.0\columnwidth]{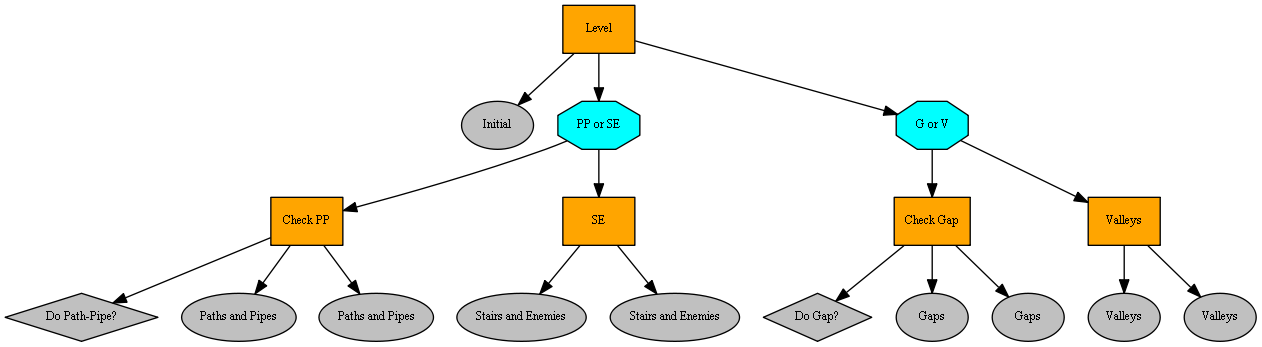}}} \\
\includegraphics[width=1\columnwidth]{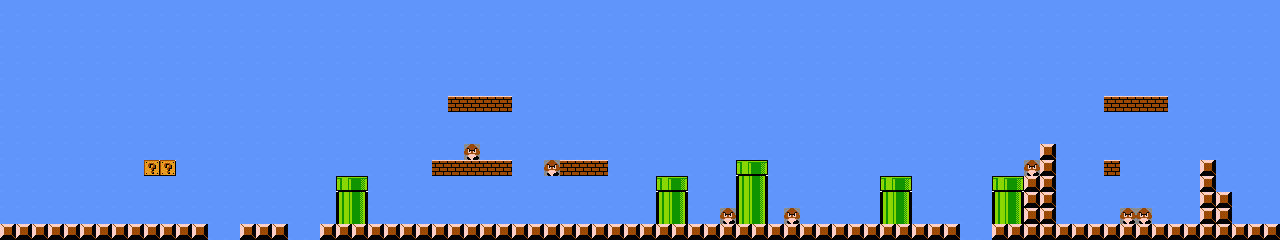}         & \includegraphics[width=1\columnwidth]{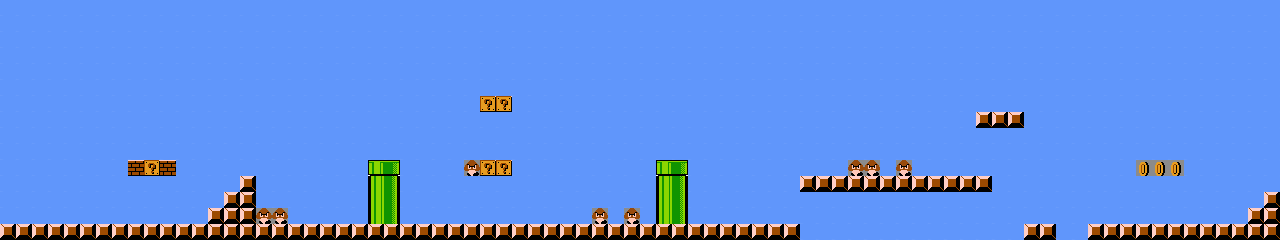}   
\end{tabular}
\caption{\label{XFIGUREsmblevelbt} SMB PCG-BT and example generated levels}
\end{figure*}
}

\newcommand{\XFIGUREsmblevelbtold}{
\begin{figure*}[t]
\centering
\begin{tabular}{c}
\includegraphics[width=1.7\columnwidth]{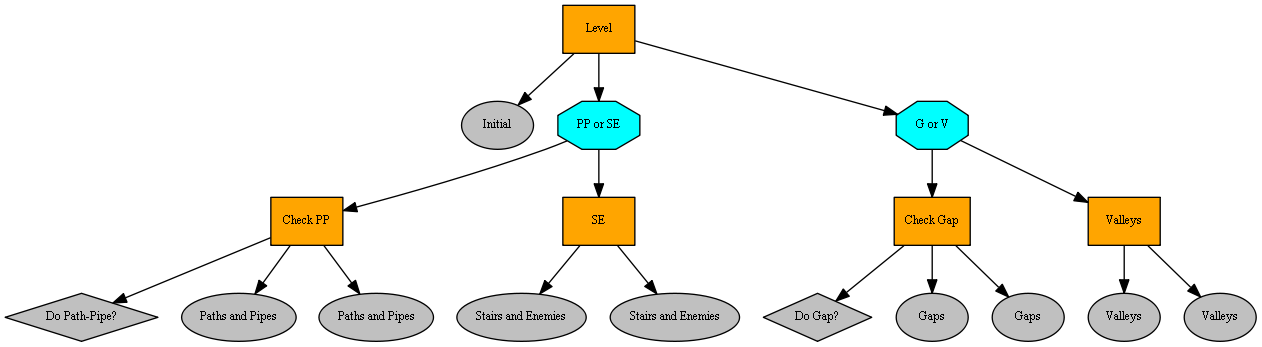}
\\
\includegraphics[width=0.5\columnwidth]{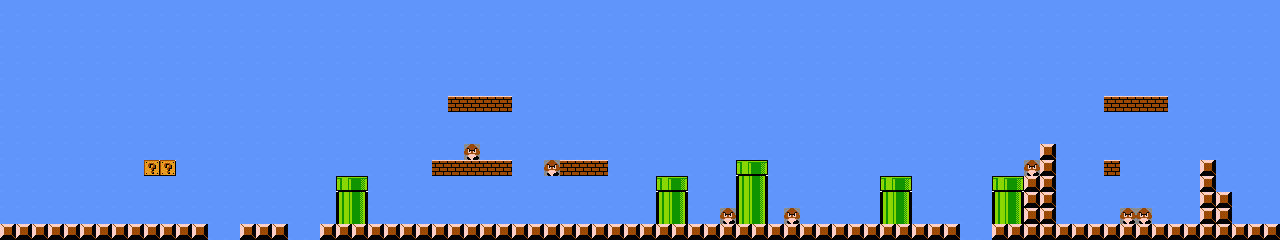}
\\
\includegraphics[width=0.5\columnwidth]{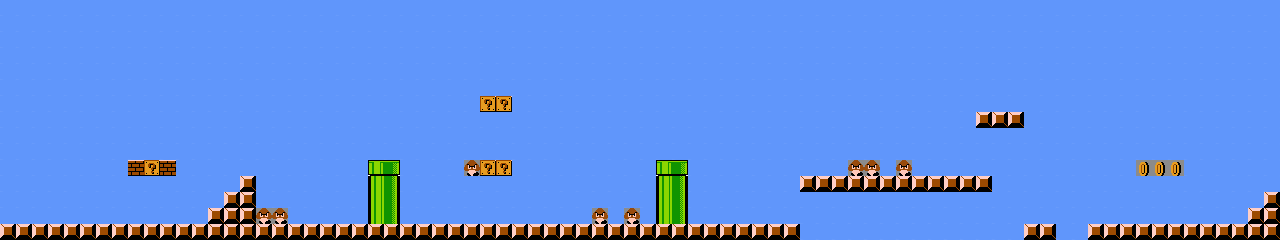}
\end{tabular}
\caption{\label{XFIGUREsmblevelbtold} SMB PCG-BT and example generated levels}
\end{figure*}
}

\newcommand{\XFIGUREmmoneone}{
\begin{figure*}[t]
\centering
\begin{tabular}{c}
\includegraphics[width=2\columnwidth]{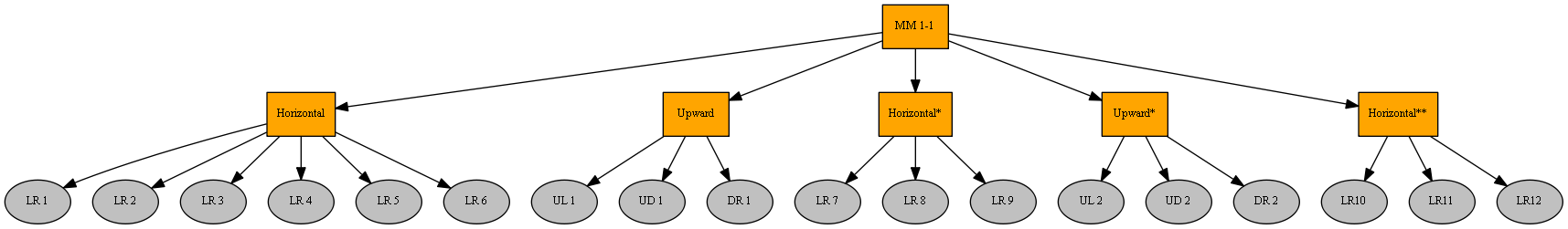}
\\
\raisebox{5pt}{\rotatebox{90}{\scriptsize{verbatim}}}
\includegraphics[width=1.5\columnwidth]{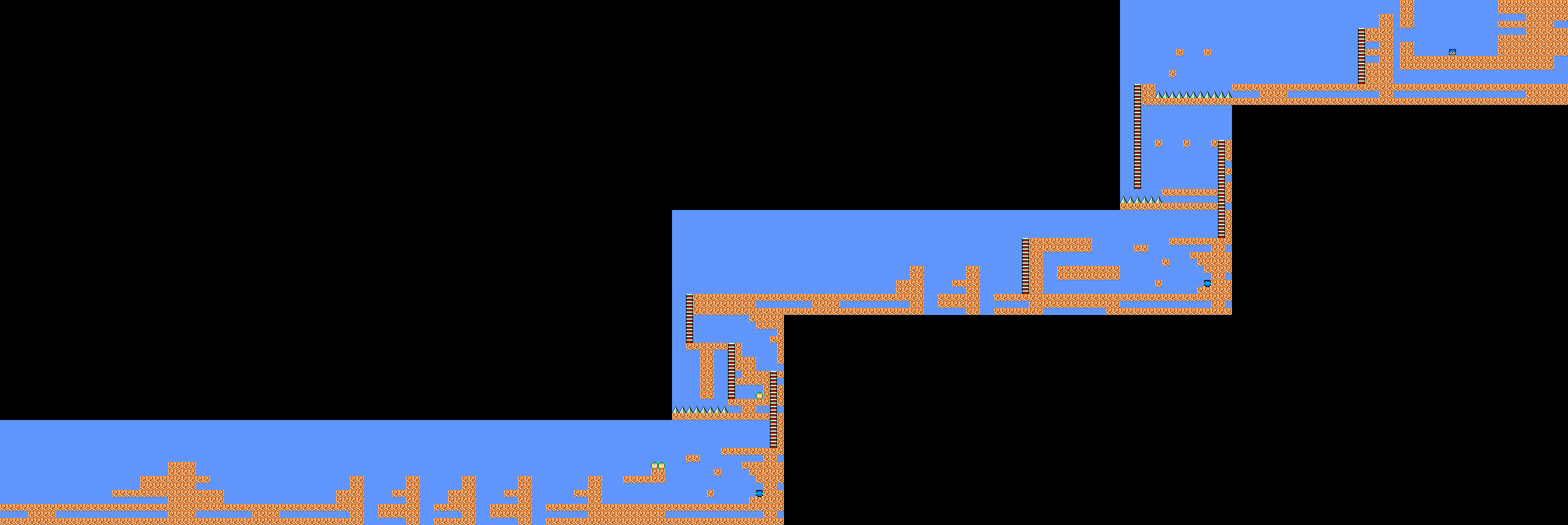}
\\
\raisebox{5pt}{\rotatebox{90}{\scriptsize{sampled}}}
\includegraphics[width=1.5\columnwidth]{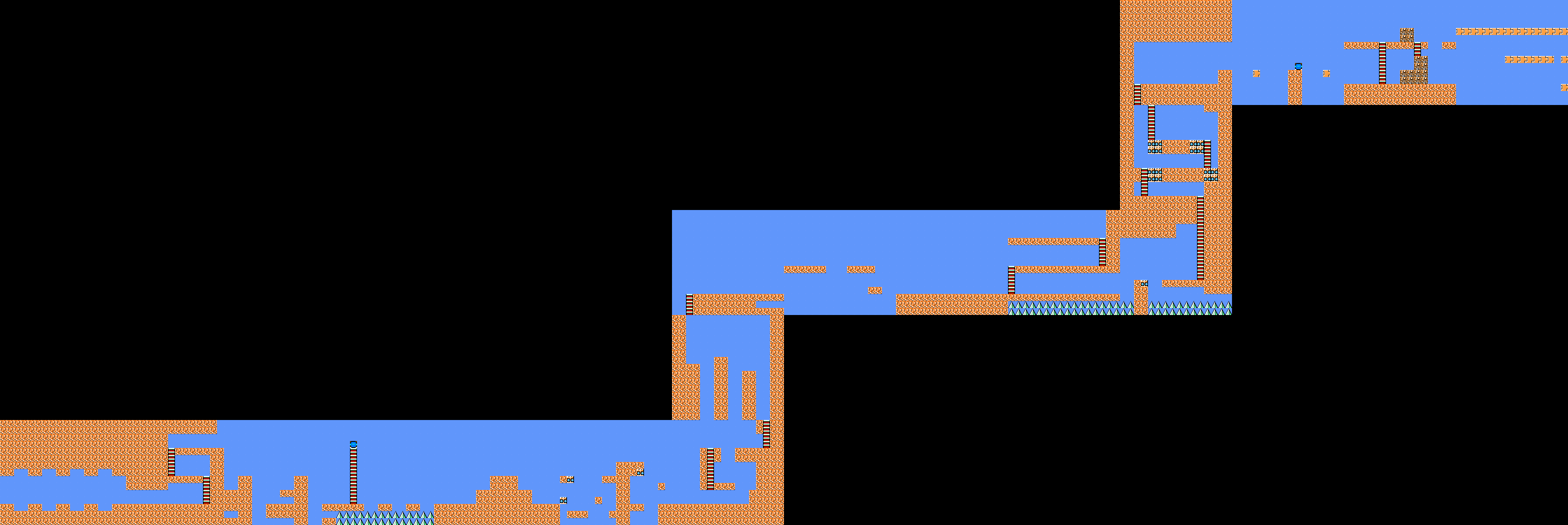}
\end{tabular}
\caption{\label{XFIGUREmmoneone} MM 1-1}
\end{figure*}
}

\newcommand{\XFIGUREmmlevelbtonecol}{
\begin{figure}[t]
\centering
\begin{tabular}{c}
{\includegraphics[width=1.0\columnwidth]{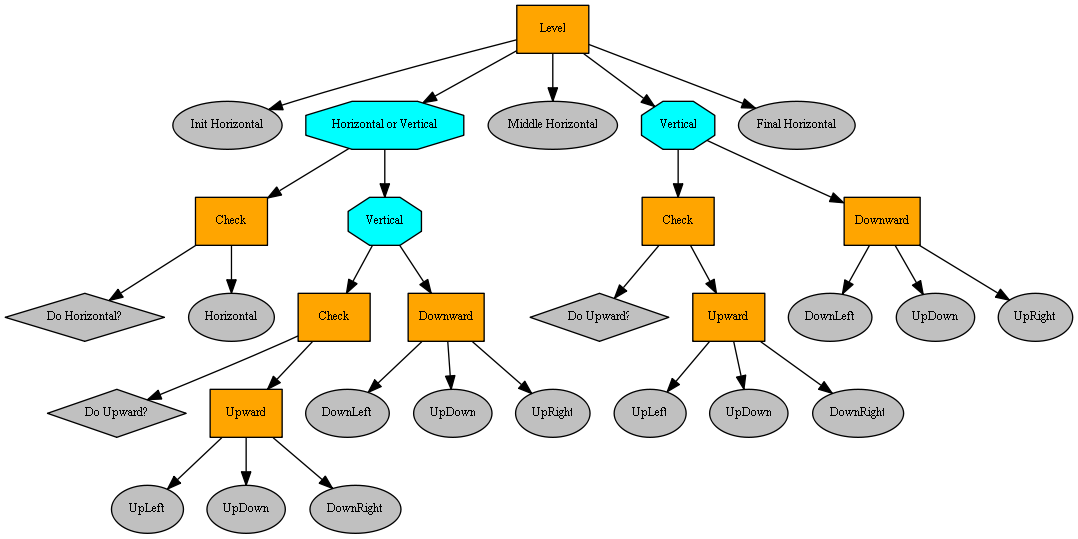}} \\
\includegraphics[width=1.0\columnwidth]{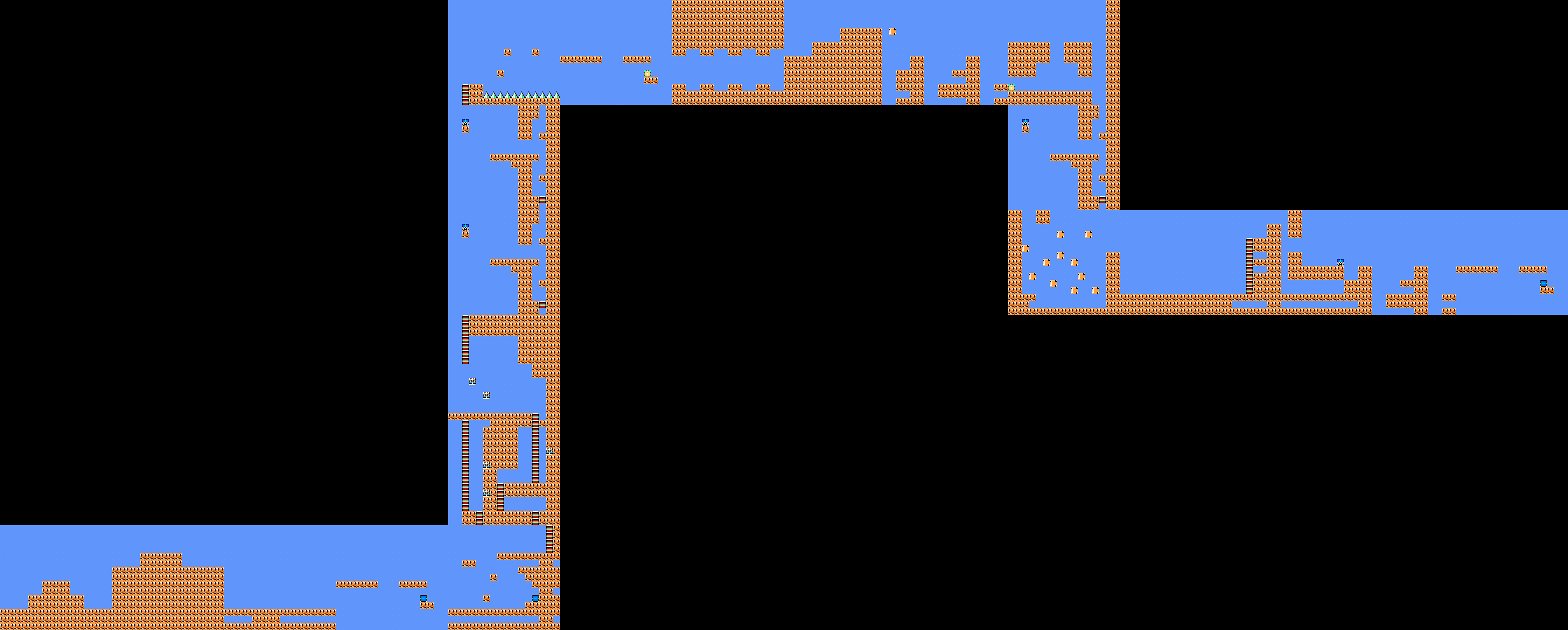}           \\ \includegraphics[width=1.0\columnwidth]{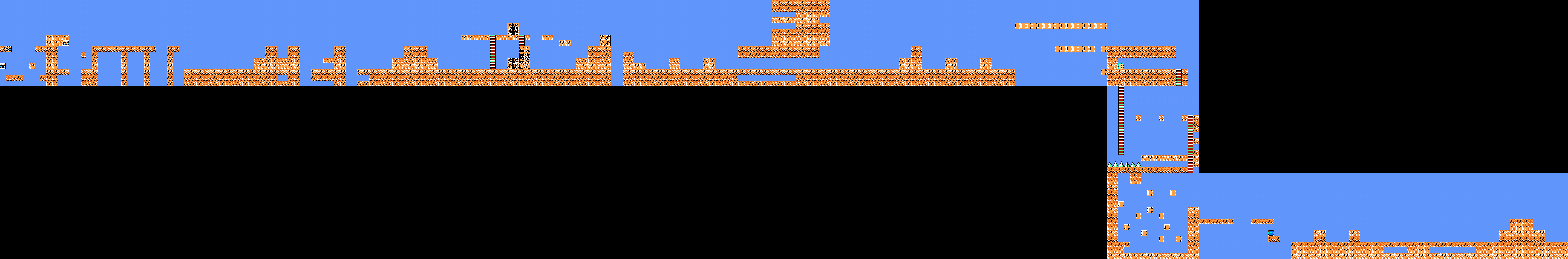}        
\end{tabular}
\caption{\label{XFIGUREmmlevelbtonecol} MM PCG-BT and example generated levels.}
\end{figure}
}

\newcommand{\XFIGUREmmlevelbt}{
\begin{figure*}[t]
\hspace*{-0.75cm}
\centering
\begin{tabular}{cc}
\multicolumn{2}{c}{{\includegraphics[width=1.75\columnwidth]{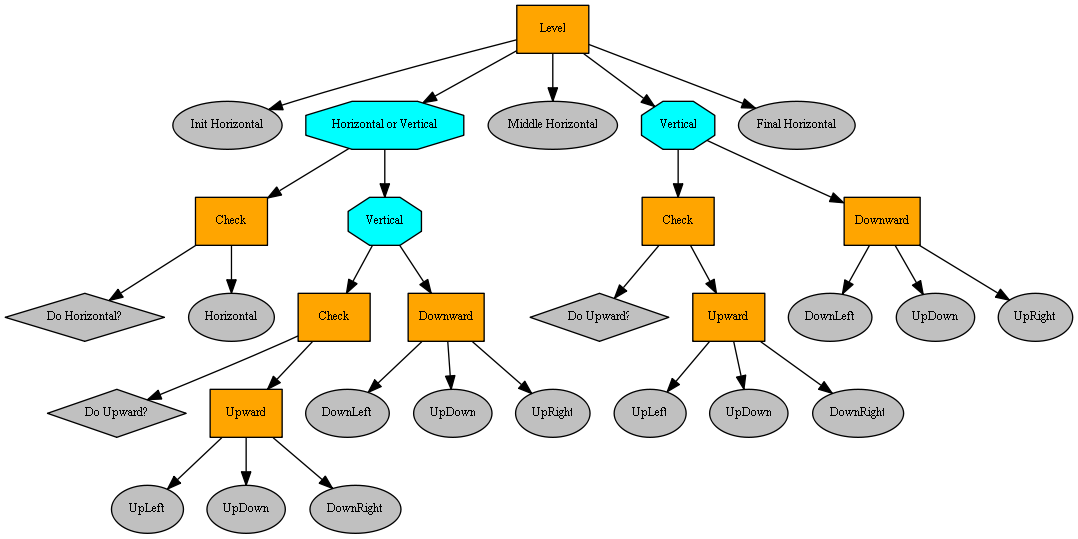}}} \\
\includegraphics[width=1\columnwidth]{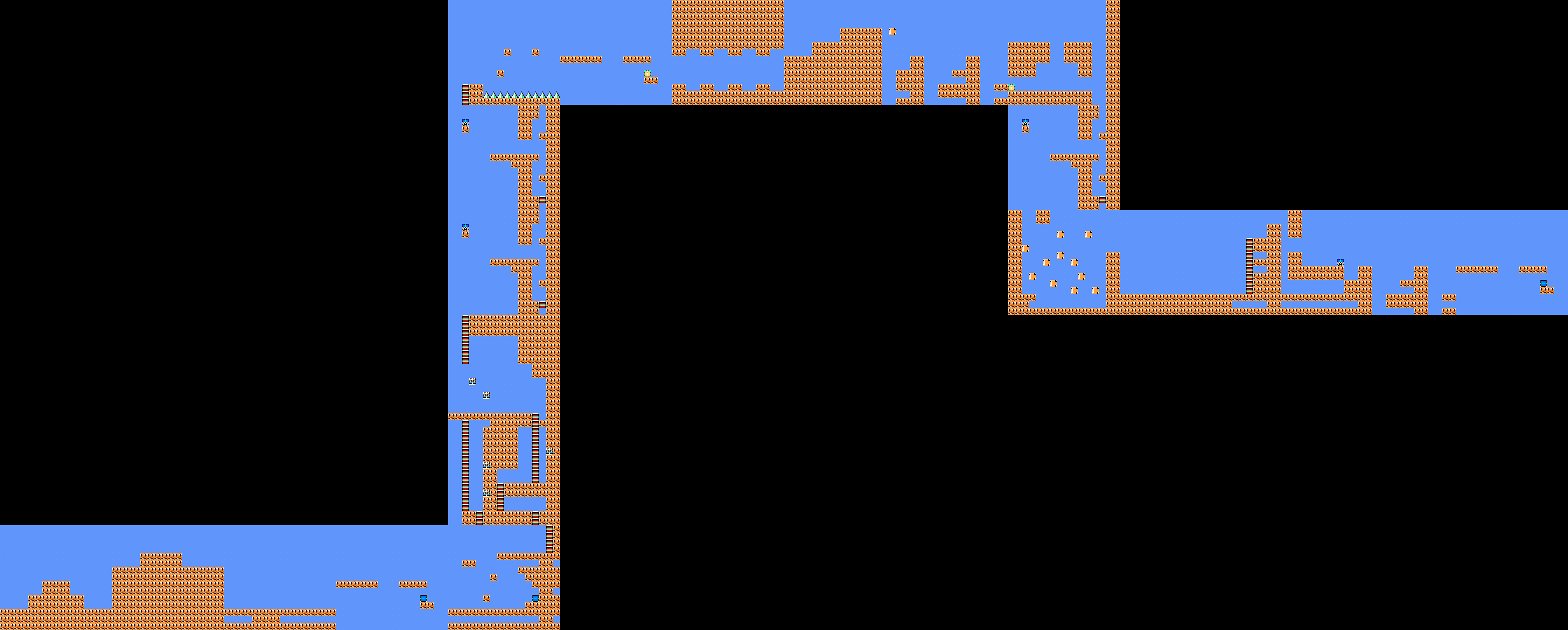}         & \includegraphics[width=1\columnwidth]{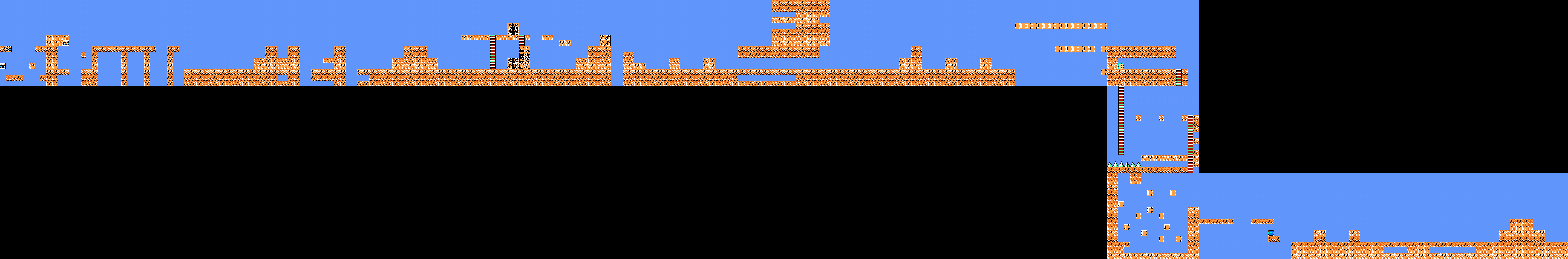}   
\end{tabular}
\caption{\label{XFIGUREmmlevelbt} MM PCG-BT and example generated levels.}
\end{figure*}
}

\newcommand{\XFIGUREmmlevelbtold}{
\begin{figure*}[t]
\centering
\begin{tabular}{c}
\includegraphics[width=1.6\columnwidth]{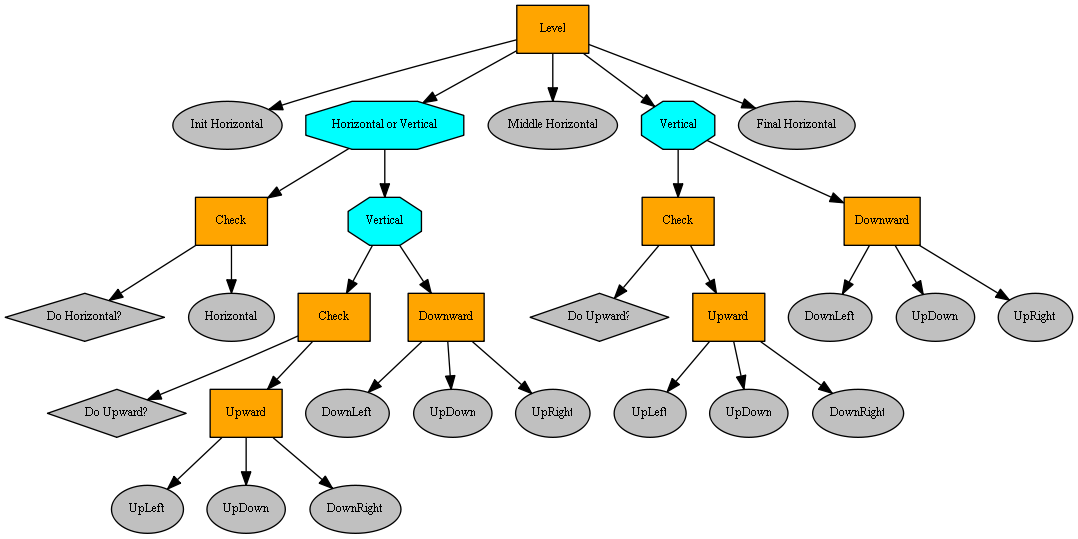}
\\
\includegraphics[width=1.25\columnwidth]{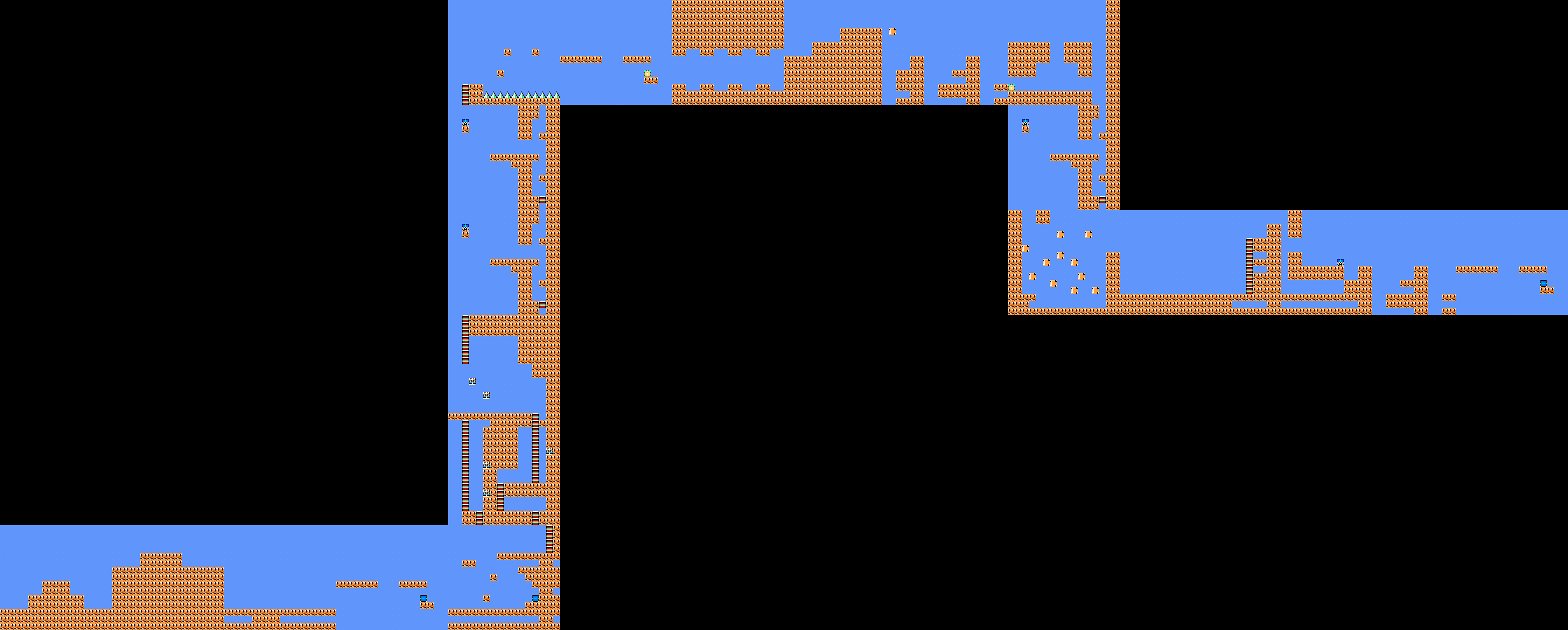}
\\
\includegraphics[width=1.5\columnwidth]{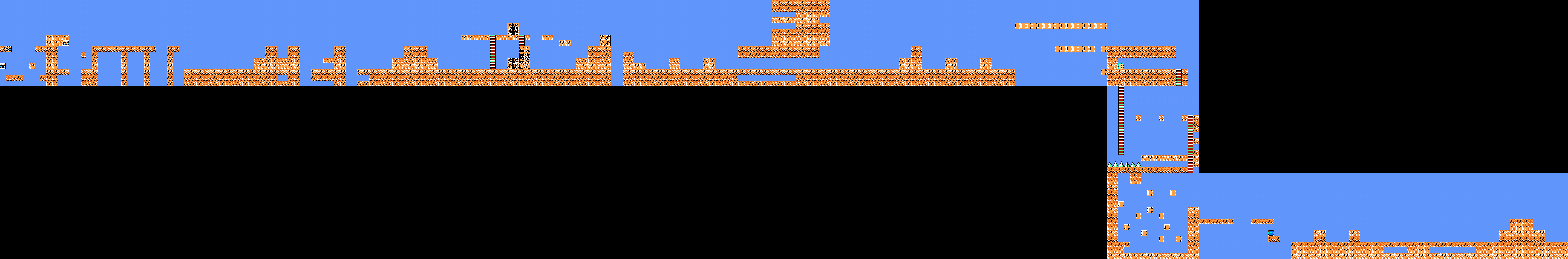}
\end{tabular}
\caption{\label{XFIGUREmmlevelbtold} MM Level}
\end{figure*}
}

\newcommand{\XFIGUREdungeonbtonecol}{
\begin{figure*}[t]
\centering
\begin{tabular}{ccc}
\includegraphics[width=0.8\columnwidth]{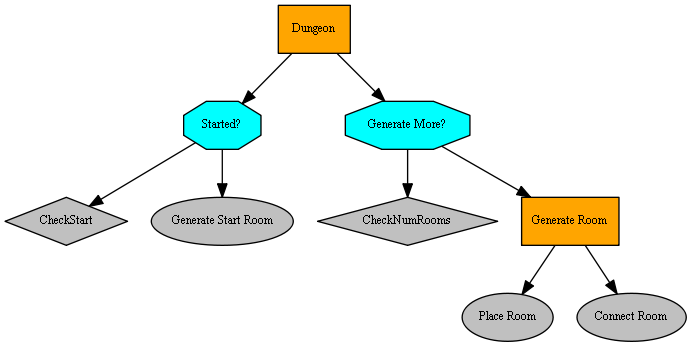}
&
\includegraphics[width=0.675\columnwidth]{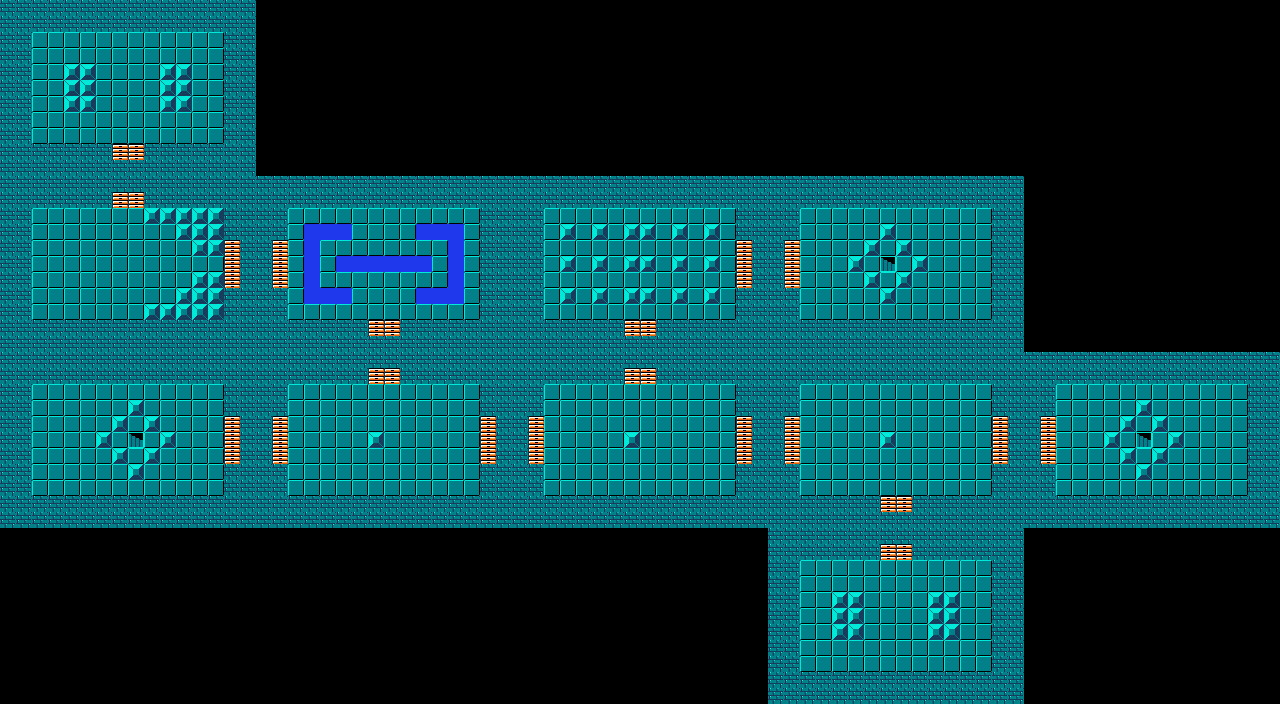}
&
\includegraphics[width=0.675\columnwidth]{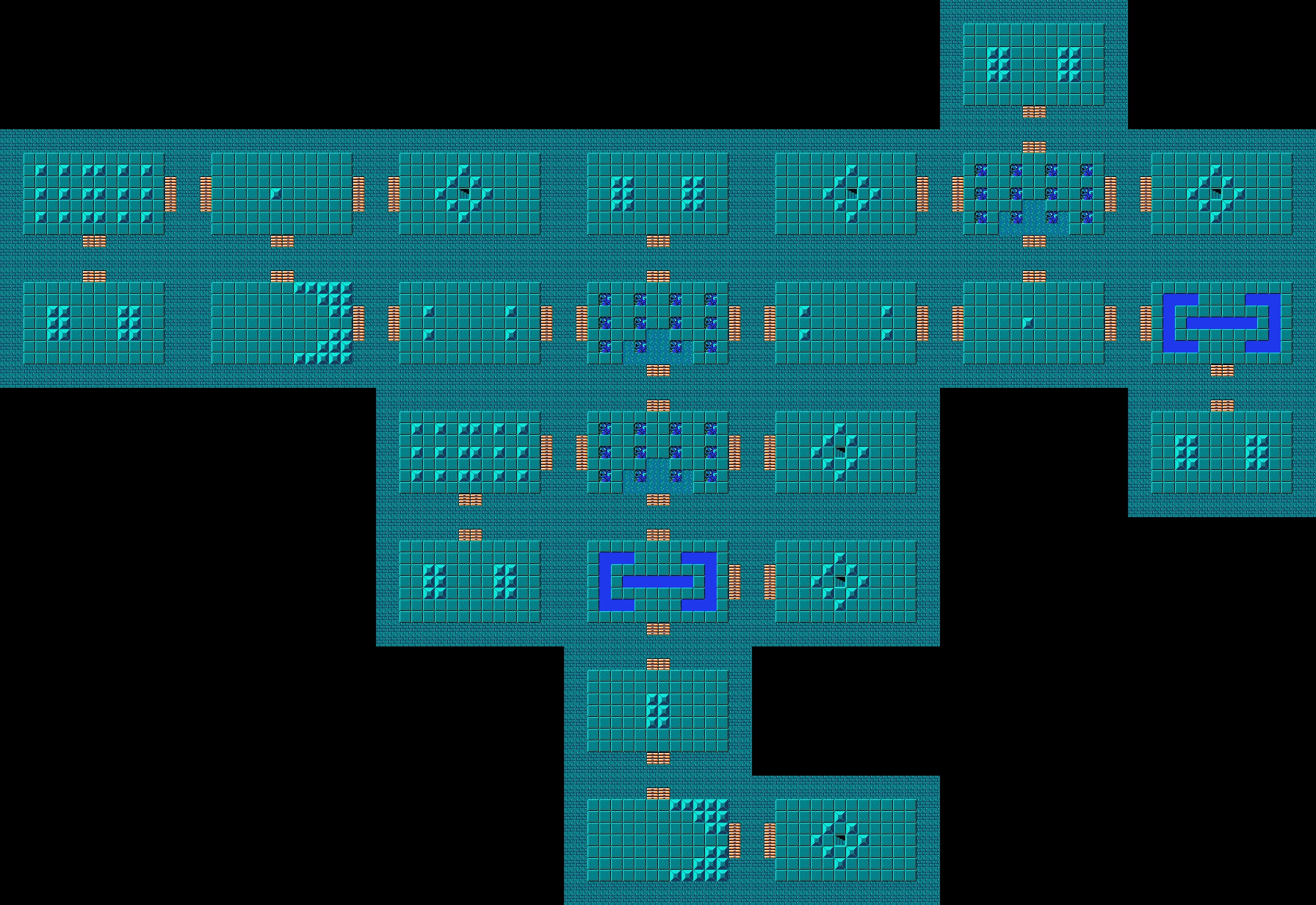}
\end{tabular}
\caption{\label{XFIGUREdungeonbtonecol} Dungeon Layout PCG-BT and example generated dungeons.}
\end{figure*}
}

\newcommand{\XFIGUREdungeonbtacross}{
\begin{figure*}[t]
\hspace*{-0.75cm}
\centering
\begin{tabular}{cc}
\multirow{3}{*}{{\includegraphics[width=1.25\columnwidth]{figure/dungeon_bt}}} &
\includegraphics[width=0.5\columnwidth]{figure/dungeon_1} \\         & \includegraphics[width=0.5\columnwidth]{figure/dungeon_2}   \\ &
\includegraphics[width=0.5\columnwidth]{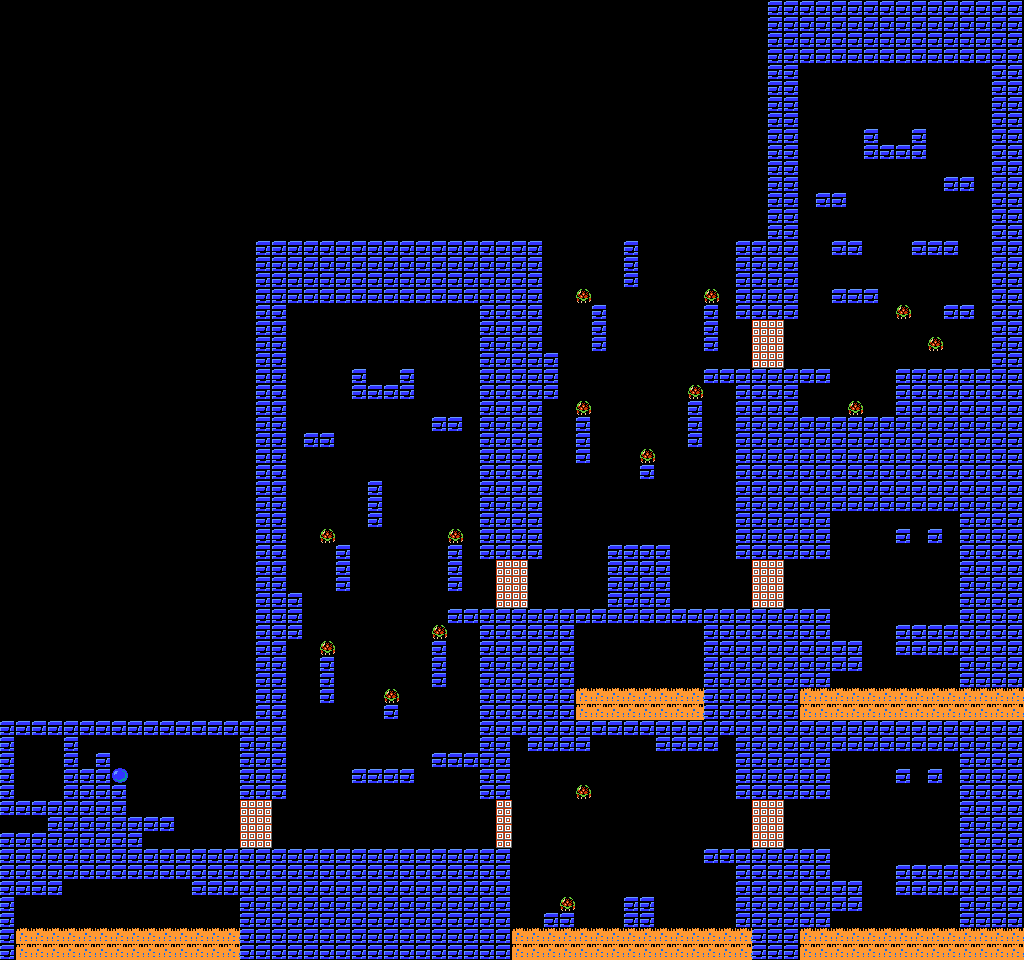}
\end{tabular}
\caption{\label{XFIGUREdungeonbtacross} Dungeon Layout PCG-BT and example generated dungeons and Metroid level}
\end{figure*}
}

\newcommand{\XFIGUREdungeonbt}{
\begin{figure}[t]
\hspace*{-0.75cm}
\centering
\begin{tabular}{c}
\includegraphics[width=0.8\columnwidth]{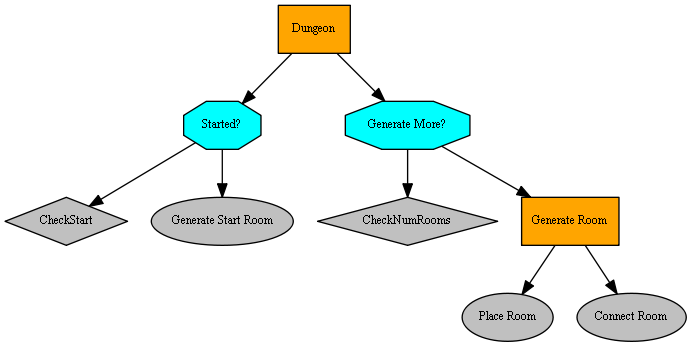} \\
\includegraphics[width=0.45\columnwidth]{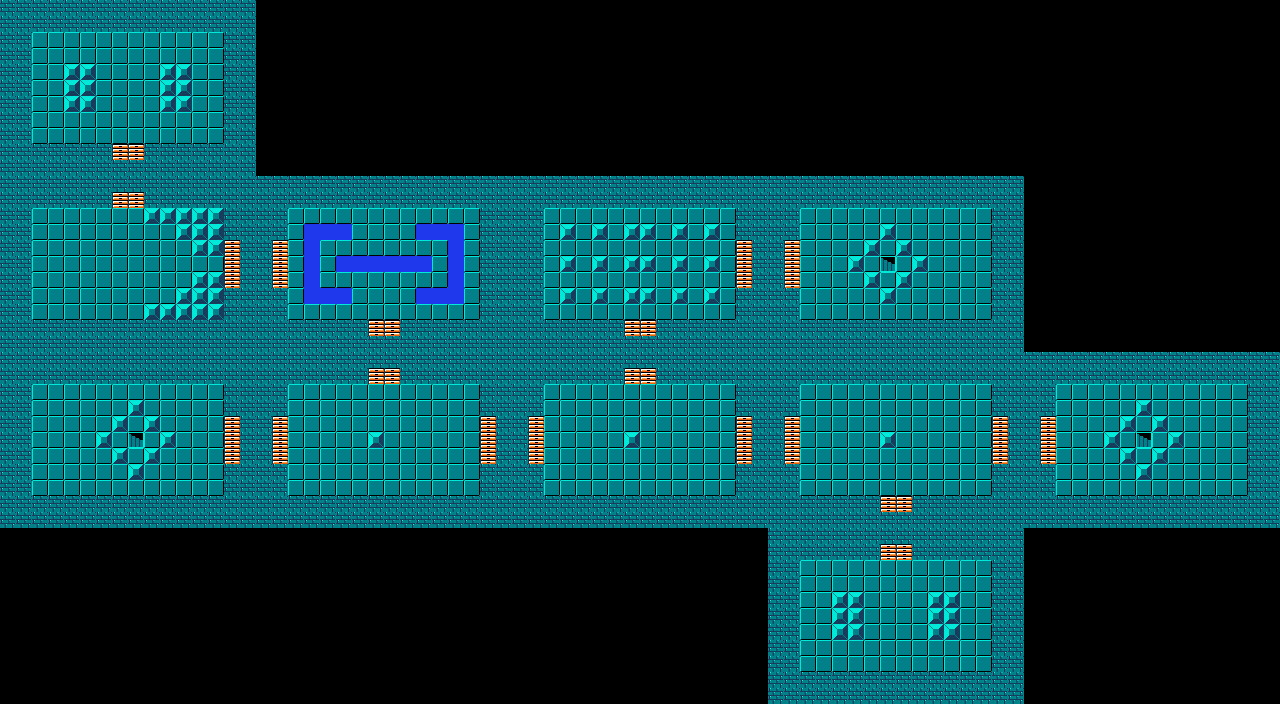} \\         
\includegraphics[width=0.45\columnwidth]{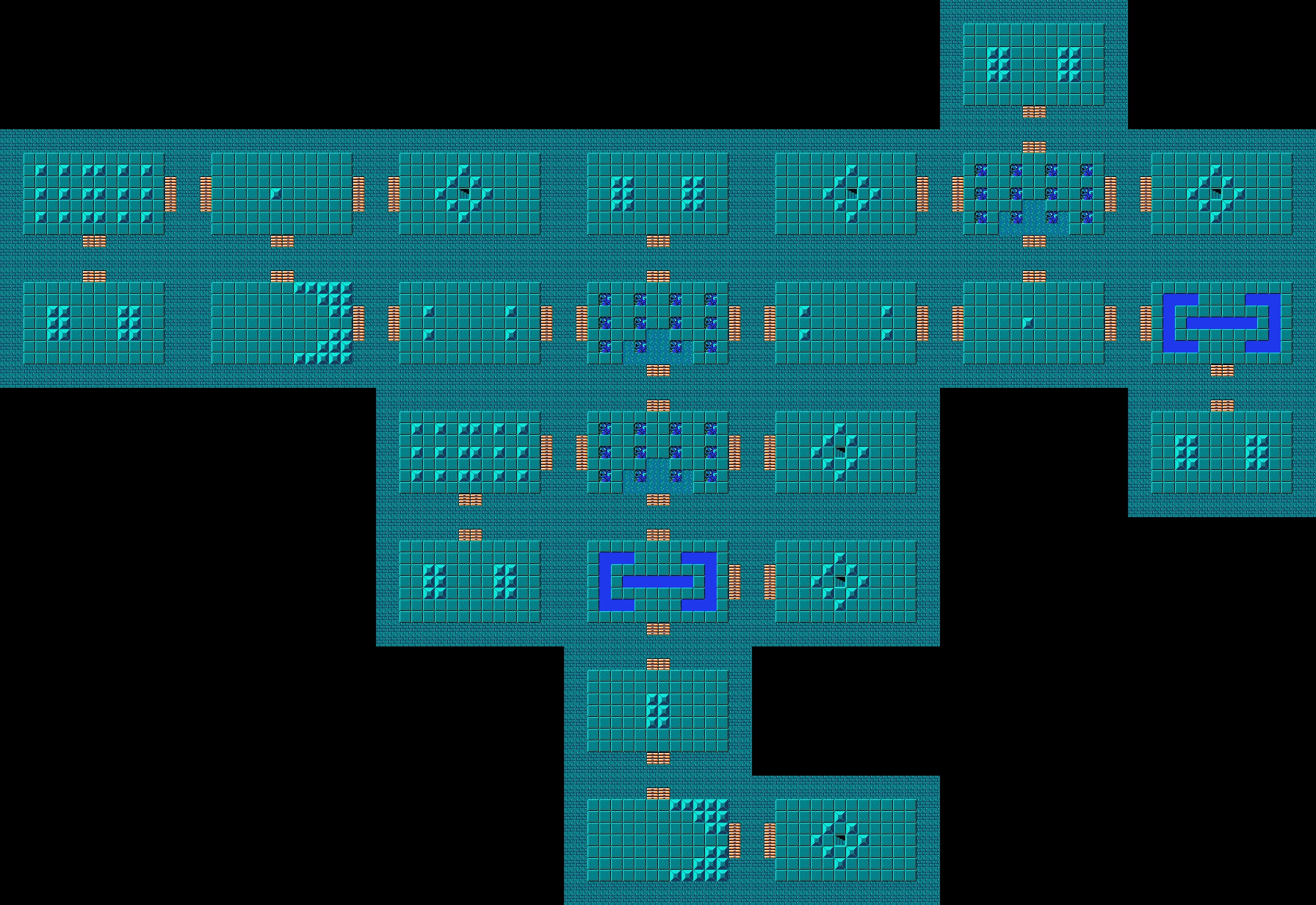}   \\ 
\includegraphics[width=0.45\columnwidth]{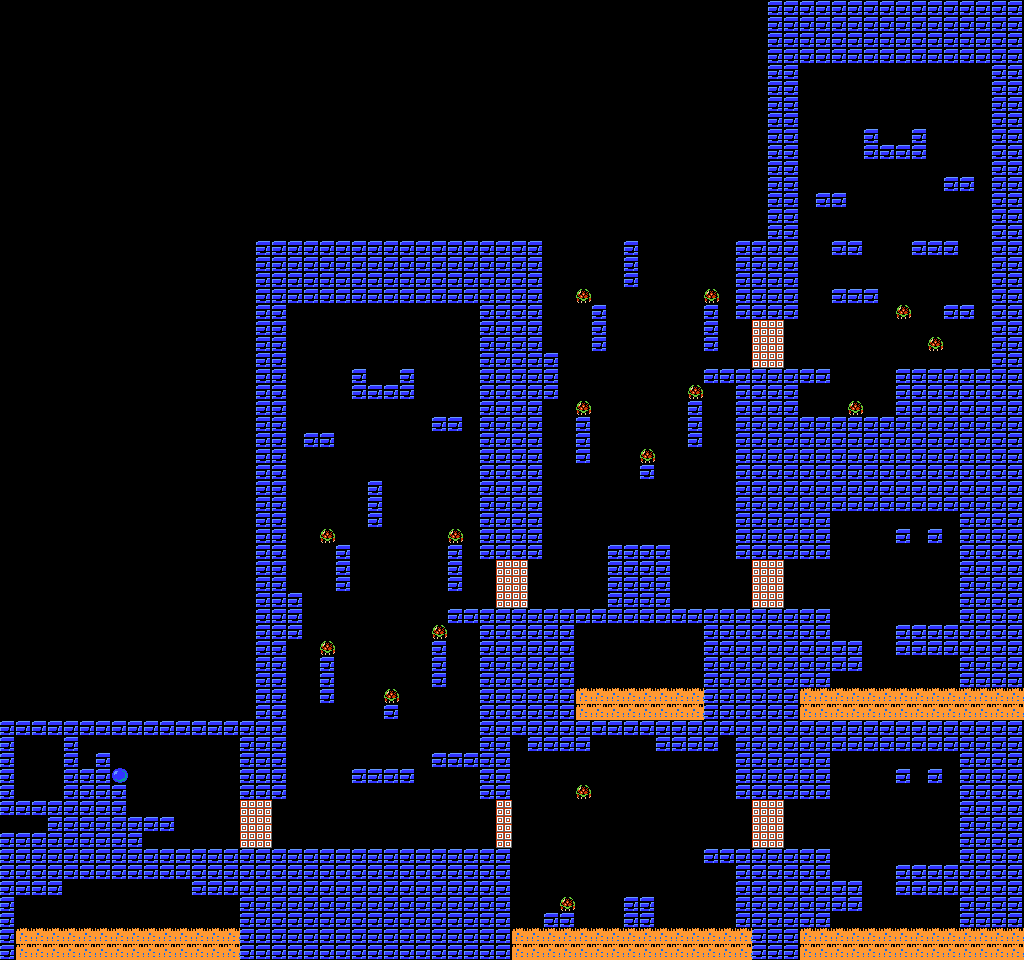}
\end{tabular}
\caption{\label{XFIGUREdungeonbt} Dungeon Layout PCG-BT and example generated dungeons and Metroid level}
\end{figure}
}

\newcommand{\XFIGUREgenericbtonecol}{
\begin{figure*}[t]
\resizebox{\textwidth}{!}{
\centering
\begin{tabular}{ccc}
\includegraphics[width=1\columnwidth]{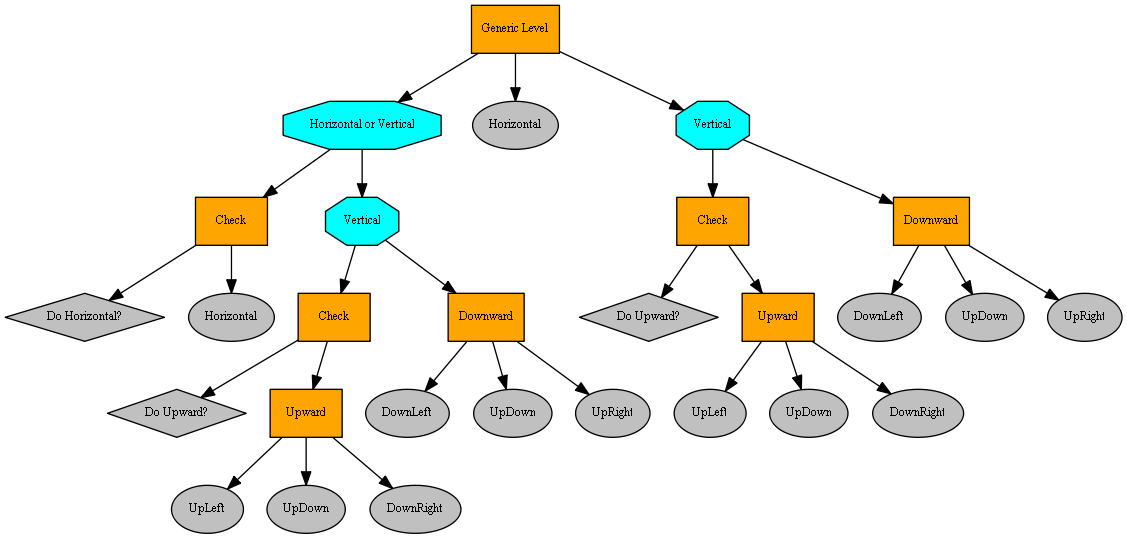}
&
\includegraphics[width=0.65\columnwidth]{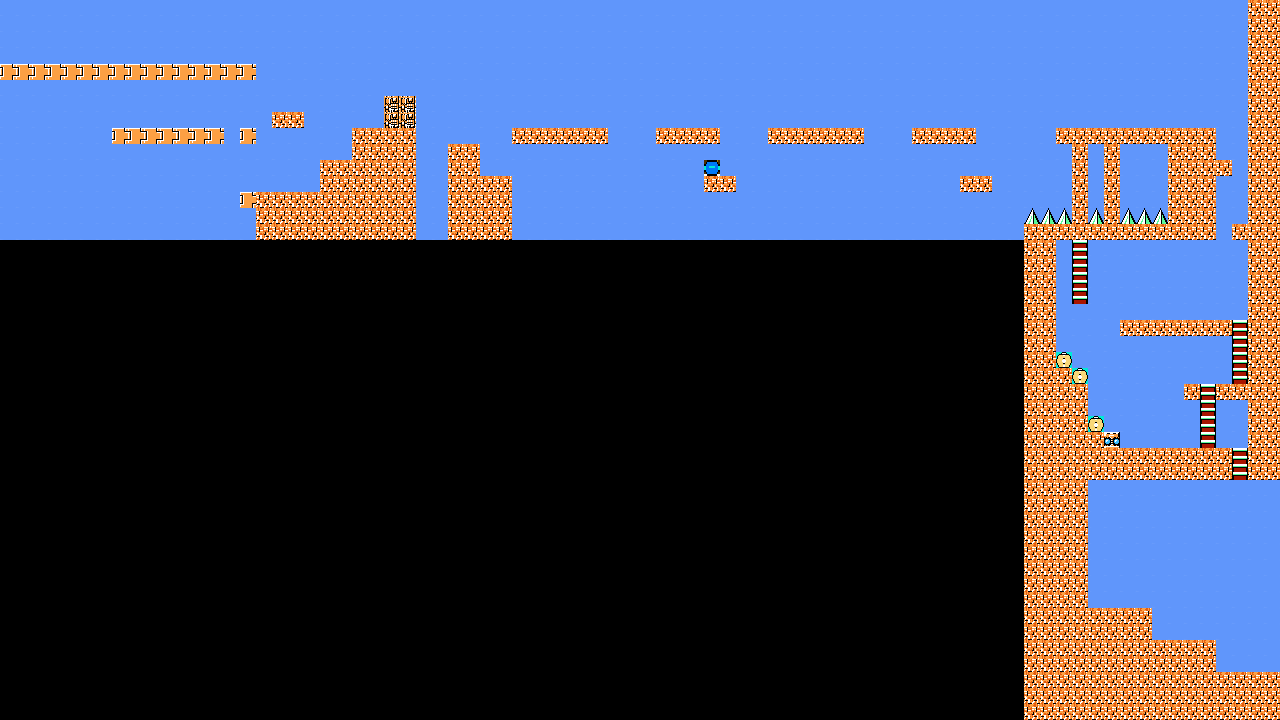}
&
\includegraphics[width=0.55\columnwidth]{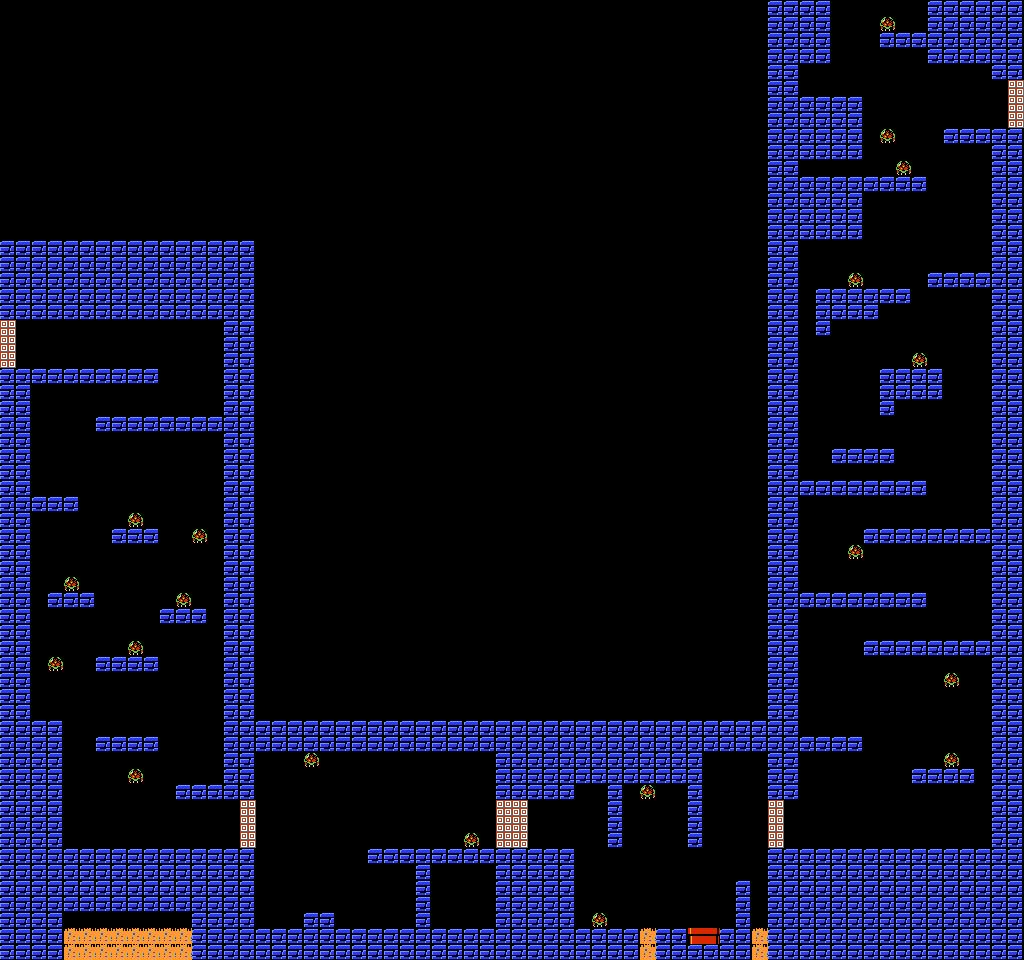}
\end{tabular}
}
\caption{\label{XFIGUREgenericbtonecol} Generic PCG-BT and example generated MM and Metroid levels.}
\end{figure*}
}

\newcommand{\XFIGUREgenericbtacross}{
\begin{figure*}[t]
\hspace*{-0.75cm}
\centering
\begin{tabular}{cc}
\multicolumn{2}{c}{{\includegraphics[width=1.75\columnwidth]{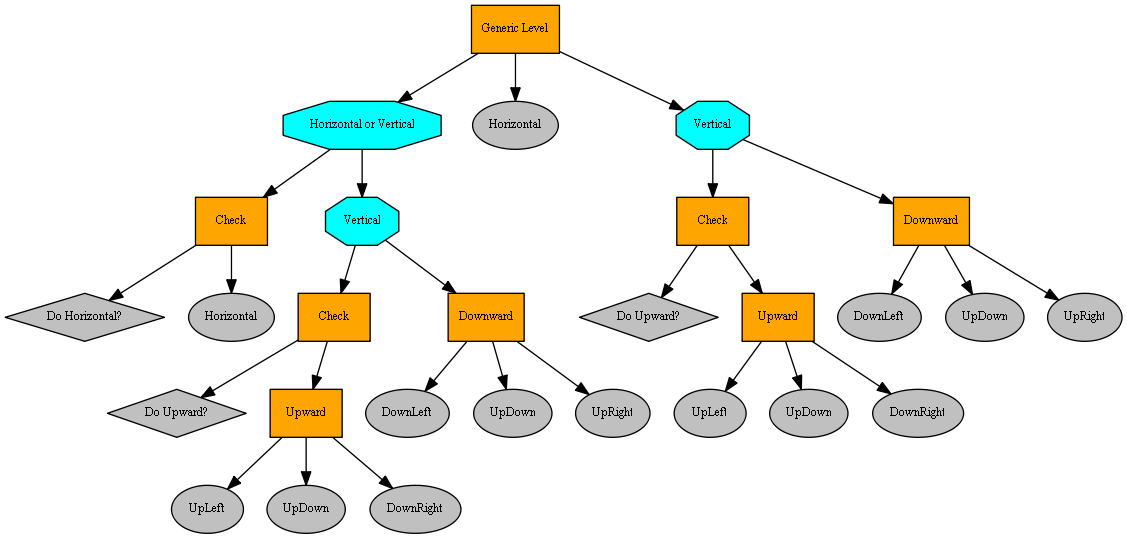}}} \\
\includegraphics[width=1\columnwidth]{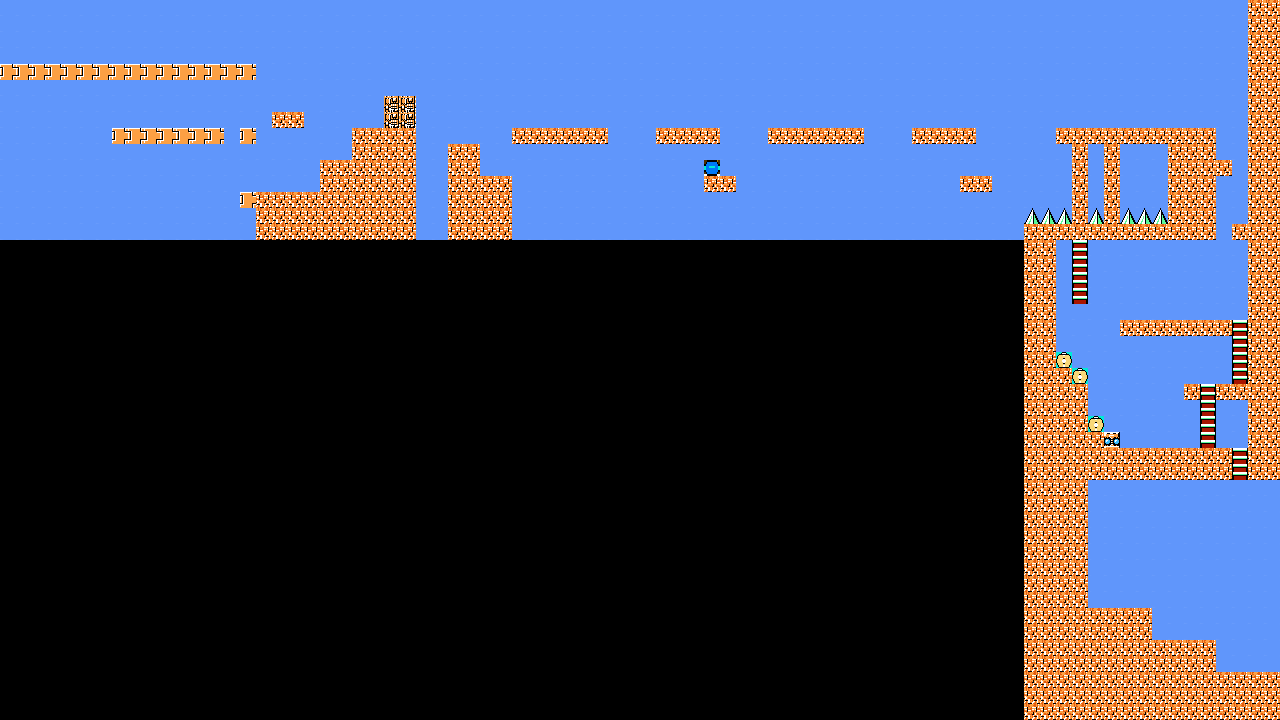}         & \includegraphics[width=1\columnwidth]{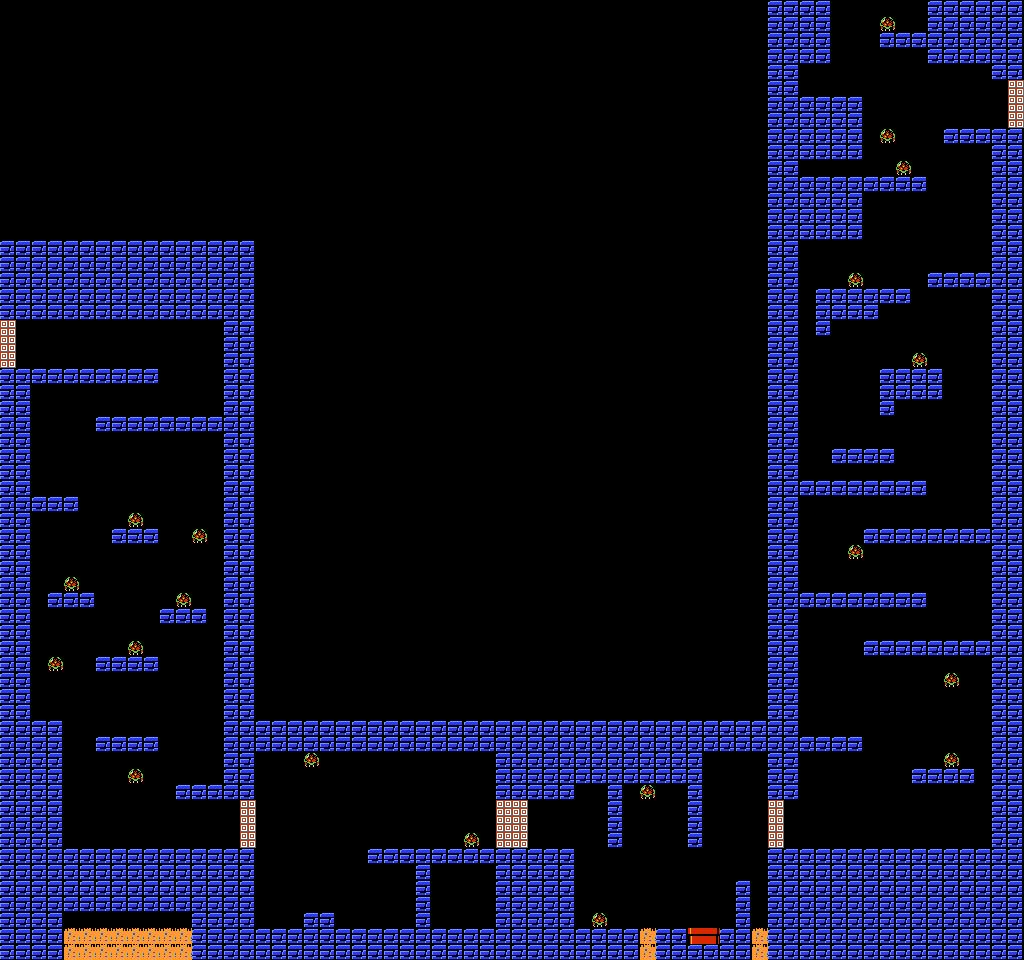}   
\end{tabular}
\caption{\label{XFIGUREgenericbtacross} Generic PCG-BT and example generated MM and Metroid levels.}
\end{figure*}
}

\newcommand{\XFIGUREgenericbt}{
\begin{figure*}[t]
\hspace*{-0.75cm}
\centering
\begin{tabular}{cc}
\multirow{2}{*}{{\includegraphics[width=1\columnwidth]{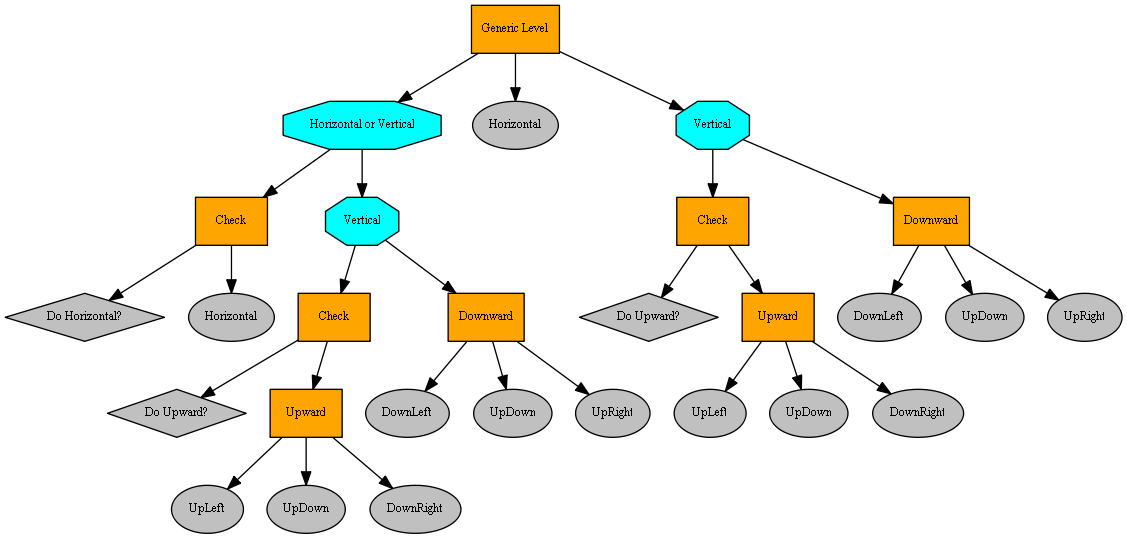}}} &
\includegraphics[width=0.5\columnwidth]{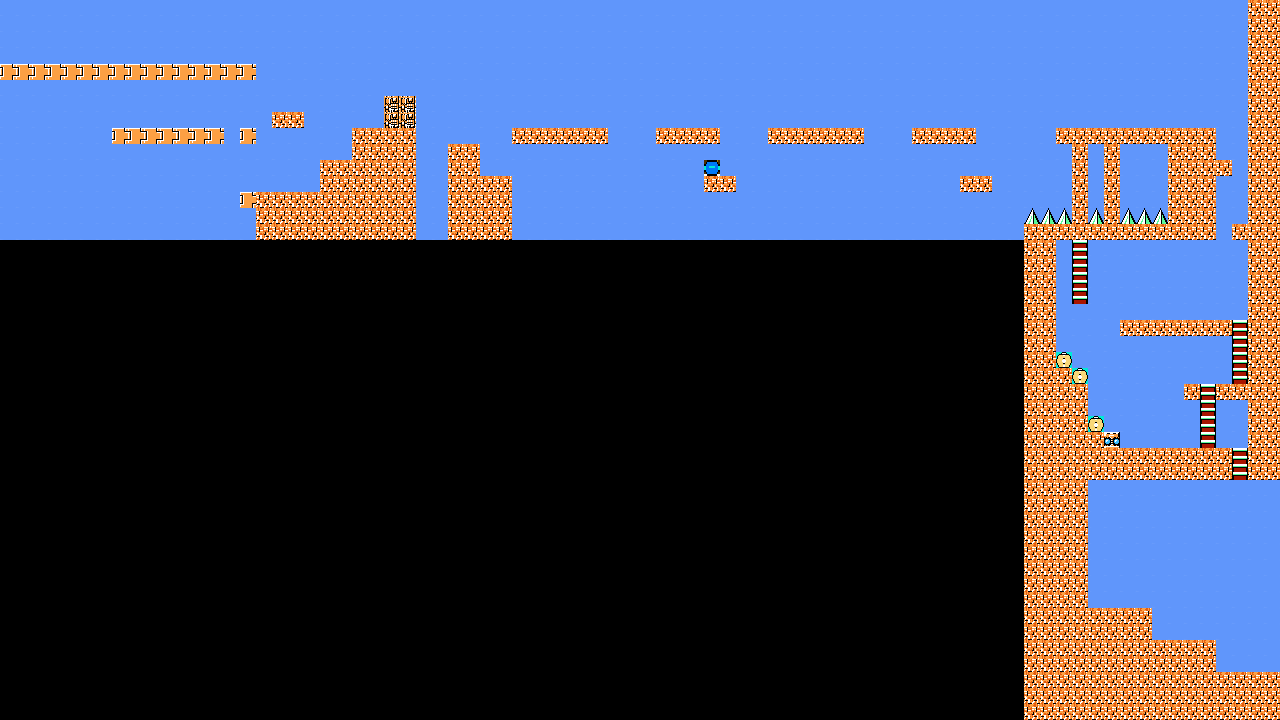}         \\ & \includegraphics[width=0.5\columnwidth]{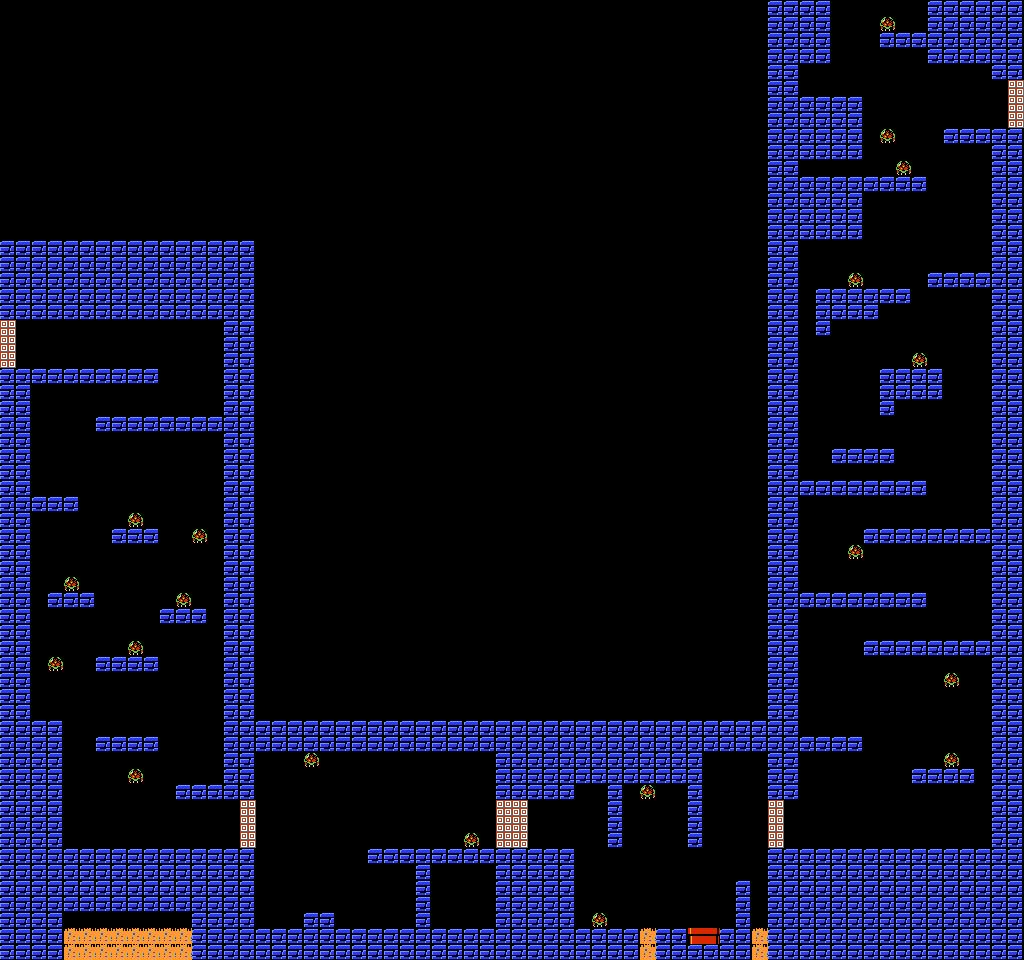}   
\end{tabular}
\caption{\label{XFIGUREgenericbt} Generic PCG-BT and example generated MM and Metroid levels.}
\end{figure*}
}

\newcommand{\XFIGUREblendbtonecol}{
\begin{figure*}[t]
\centering
\resizebox{\textwidth}{!}{
\begin{tabular}{ccc}
\includegraphics[width=1\columnwidth]{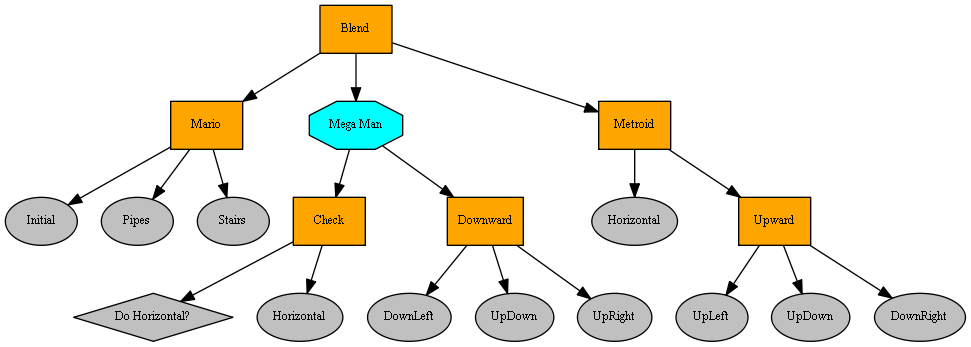}
&
\includegraphics[width=0.675\columnwidth]{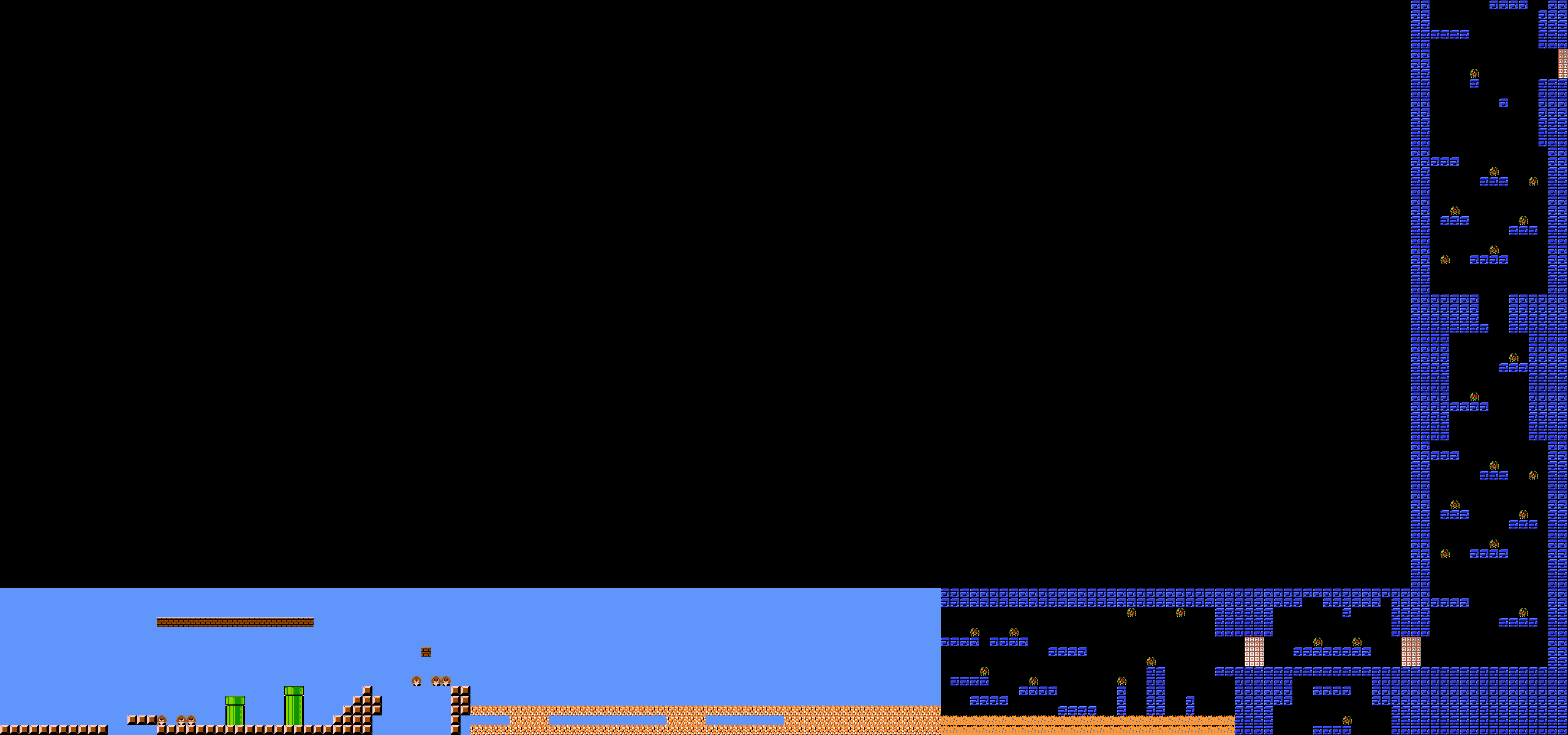}
&
\includegraphics[width=0.65\columnwidth]{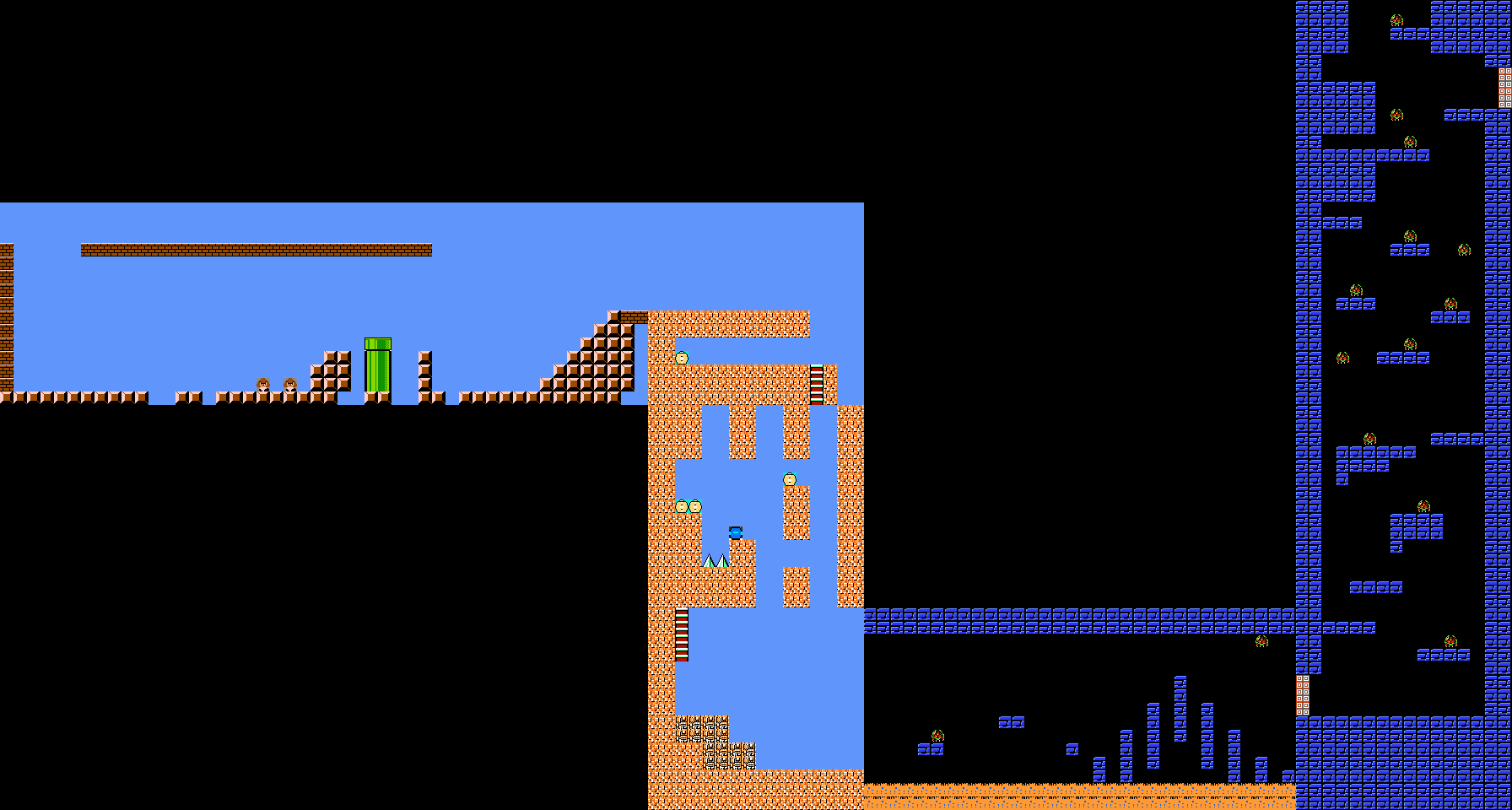}
\end{tabular}
}
\caption{\label{XFIGUREblendbtonecol} Blend PCG-BT and example generated levels blending Super Mario Bros., Mega Man and Metroid.}
\end{figure*}
}

\newcommand{\XFIGUREblendbt}{
\begin{figure*}[t]
\hspace*{-0.75cm}
\centering
\begin{tabular}{cc}
\multicolumn{2}{c}{{\includegraphics[width=2\columnwidth]{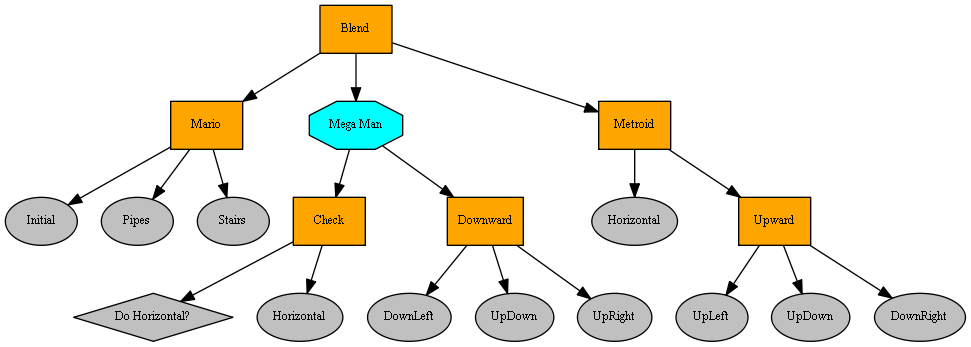}}} \\
\includegraphics[width=1\columnwidth]{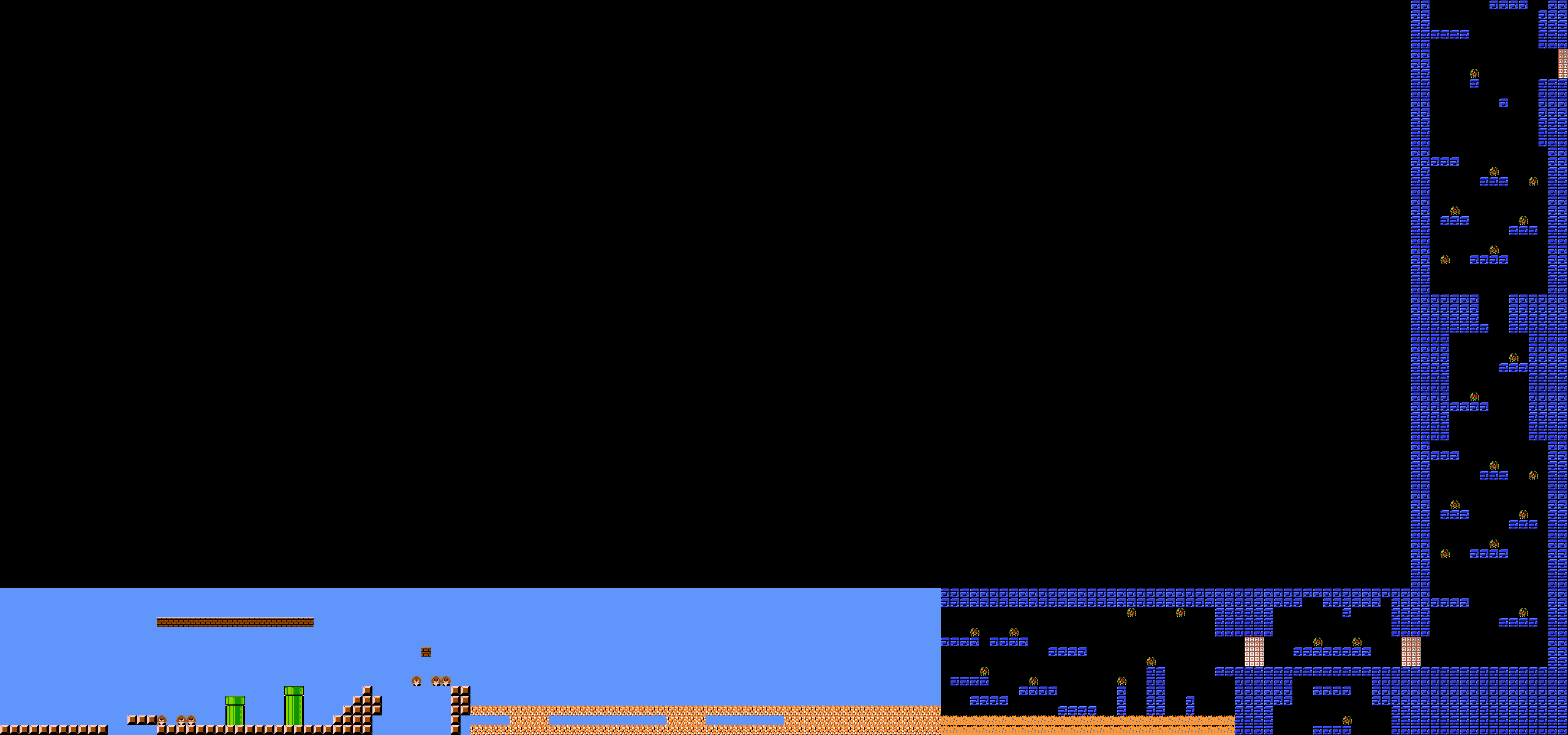}         & \includegraphics[width=1\columnwidth]{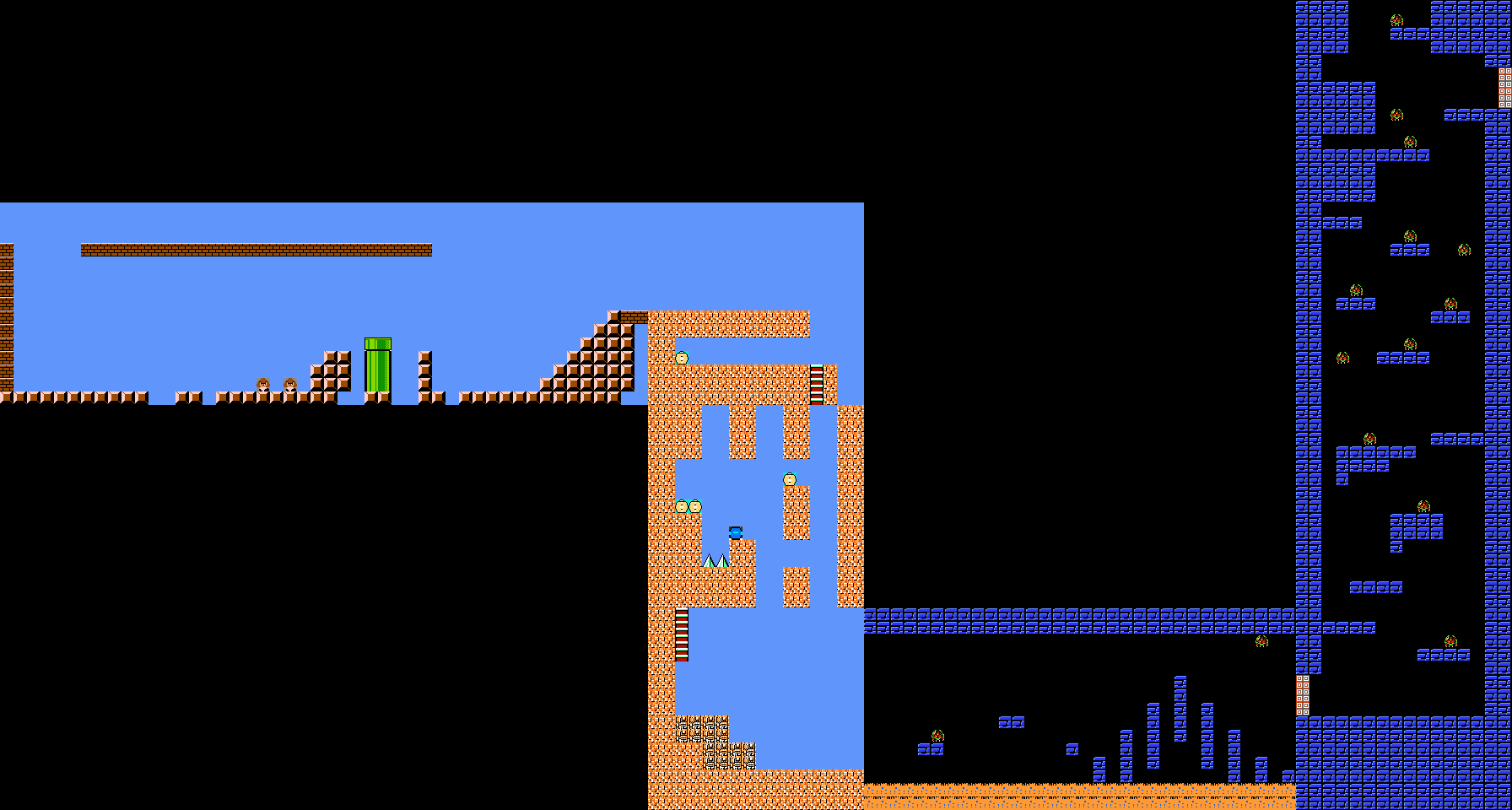}   
\end{tabular}
\caption{\label{XFIGUREblendbt} Blend PCG-BT and example generated levels blending Super Mario Bros., Mega Man and Metroid.}
\end{figure*}
}

\newcommand{\XFIGUREdungmet}{
\begin{figure}[t!]
\centering
\includegraphics[width=0.5\columnwidth]{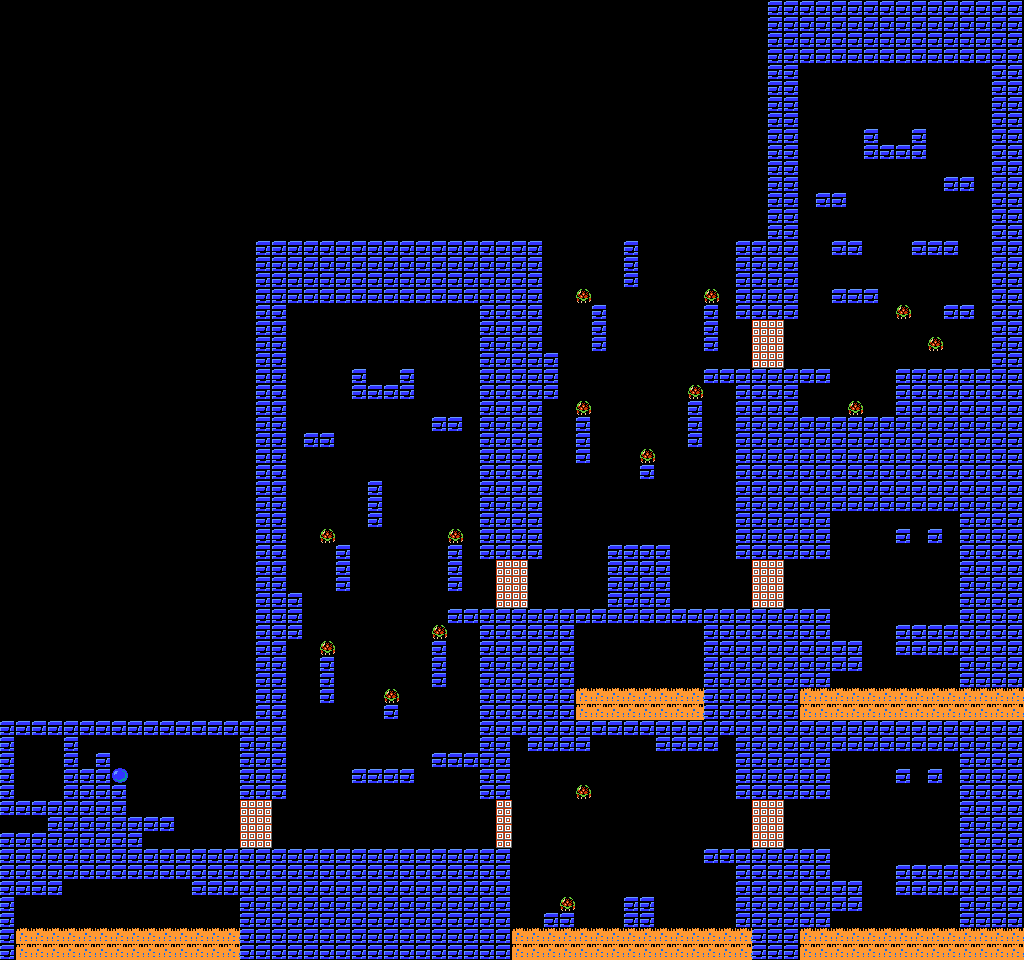}
\caption{\label{XFIGUREdungmet} Metroid level generated using dungeon layout BT.}
\end{figure}
}

\newcommand{\XFIGUREsmbworkedfix}{
\begin{figure*}
\resizebox{\textwidth}{!}{
\begin{tabular}{|cc|cc|cc|cc|}
\hline
 \includegraphics[width=.05\textwidth]{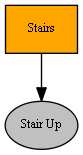}         &  \includegraphics[width=.05\textwidth]{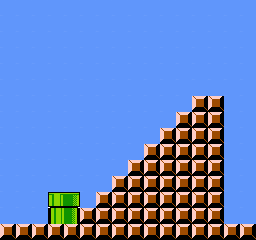}        &  \includegraphics[width=.1\textwidth]{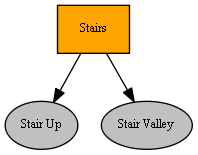}     &  
 \includegraphics[width=.1\textwidth]{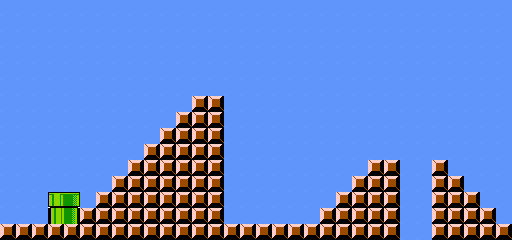}
 &
 \includegraphics[width=.05\textwidth]{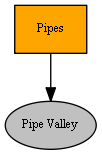}    &  \includegraphics[width=.075\textwidth]{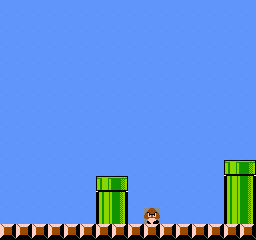}    &  \includegraphics[width=.2\textwidth]{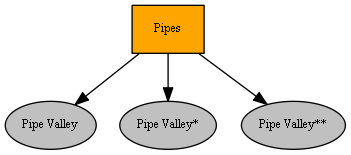}     & \includegraphics[width=.2\textwidth]{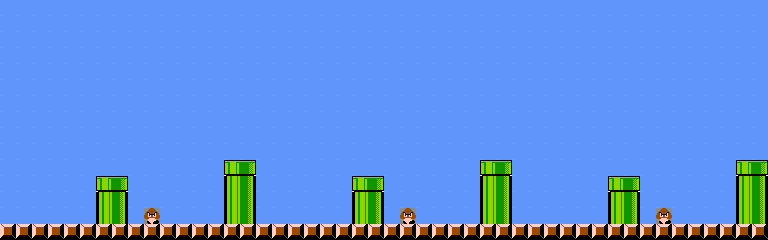}     \\
 \hline
\multicolumn{2}{|c|}{\includegraphics[width=.2\textwidth]{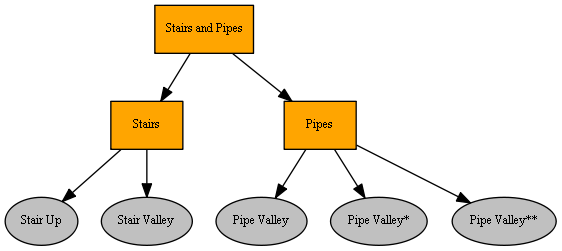} } & \multicolumn{3}{c|}{\includegraphics[width=.3\textwidth]{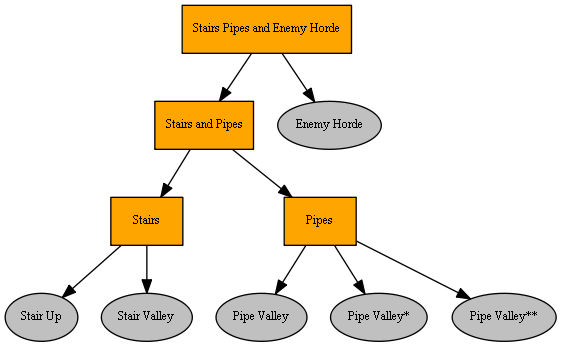}} & \multicolumn{3}{c|}{\includegraphics[width=.35\textwidth]{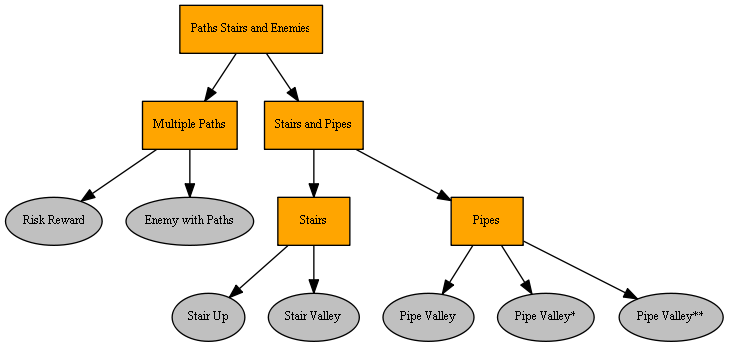}} \\
\multicolumn{2}{|c|}{\includegraphics[width=.2\textwidth]{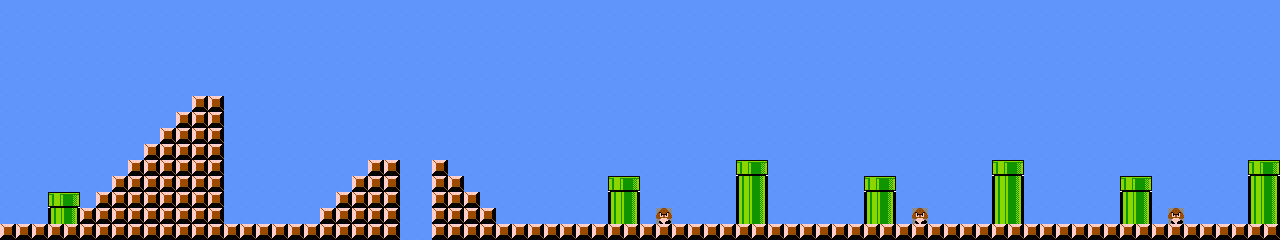}} & \multicolumn{3}{c|}{\includegraphics[width=.3\textwidth]{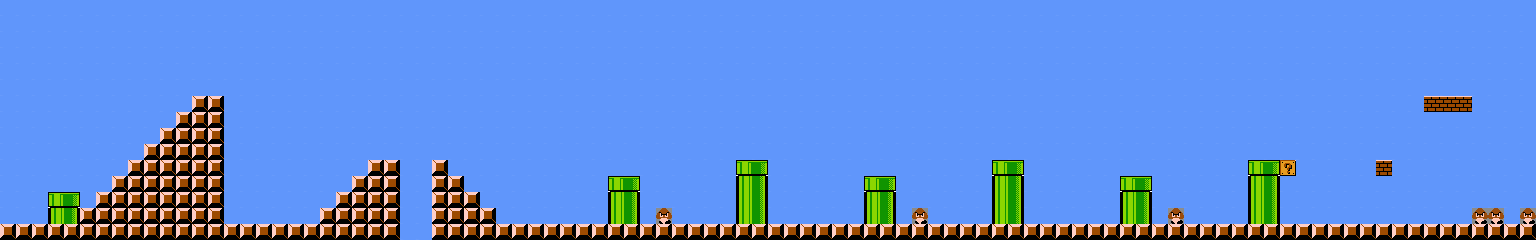}} & \multicolumn{3}{c|}{\includegraphics[width=.35\textwidth]{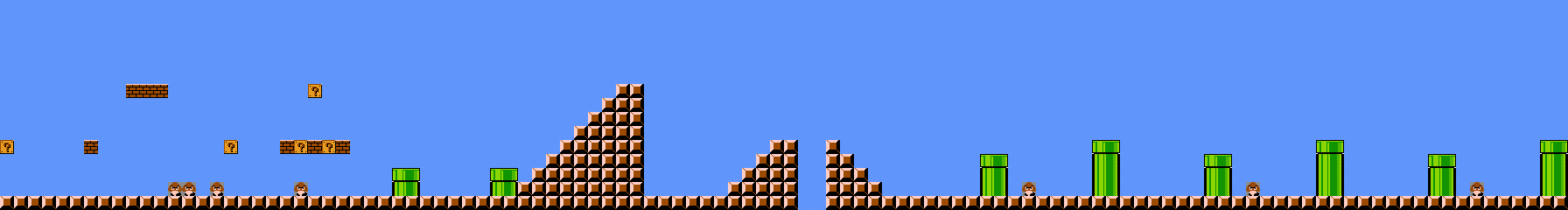}} \\
\hline
\end{tabular}
}
\caption{\label{XFIGUREsmbworkedfix} Worked example depicting a behavior tree modeling the process for generating a Mario level.}
\end{figure*}
}

\newcommand{\XFIGUREsmbworked}{
\begin{figure}
\centering
\begin{tabular}{c c}
\includegraphics[width=.05\textwidth]{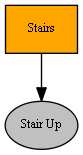} & 
\includegraphics[width=.05\textwidth]{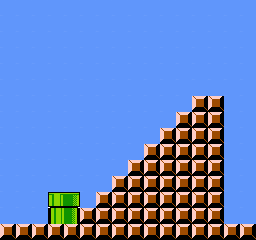} \\
\includegraphics[width=.1\textwidth]{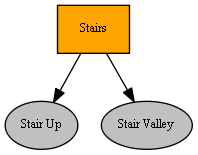} & 
\includegraphics[width=.1\textwidth]{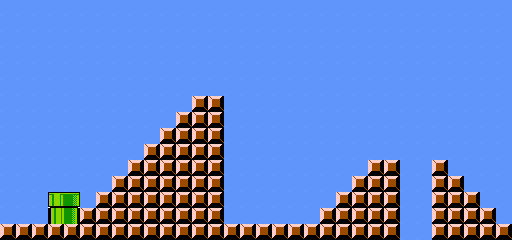} \\ 
\includegraphics[width=.05\textwidth]{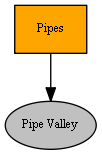} & 
\includegraphics[width=.075\textwidth]{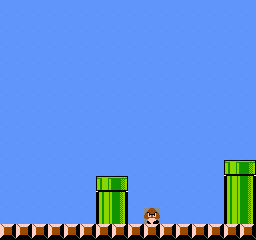} \\
\includegraphics[width=.2\textwidth]{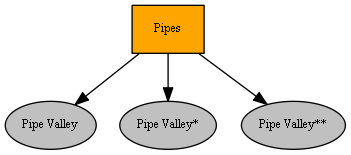} & 
\includegraphics[width=.2\textwidth]{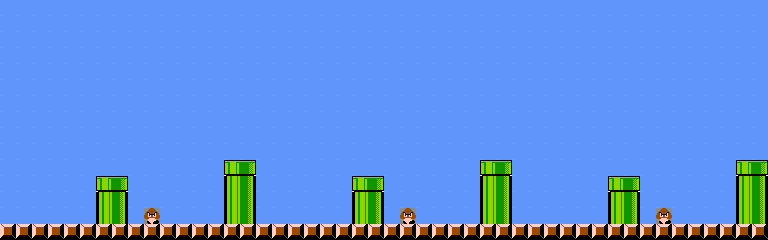} \\
\includegraphics[width=.2\textwidth]{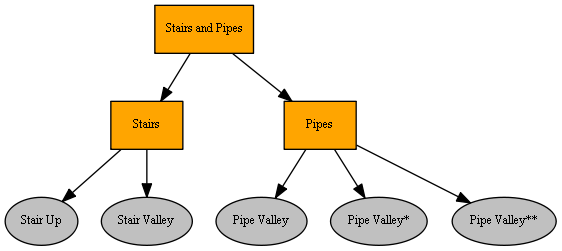} &
\includegraphics[width=.2\textwidth]{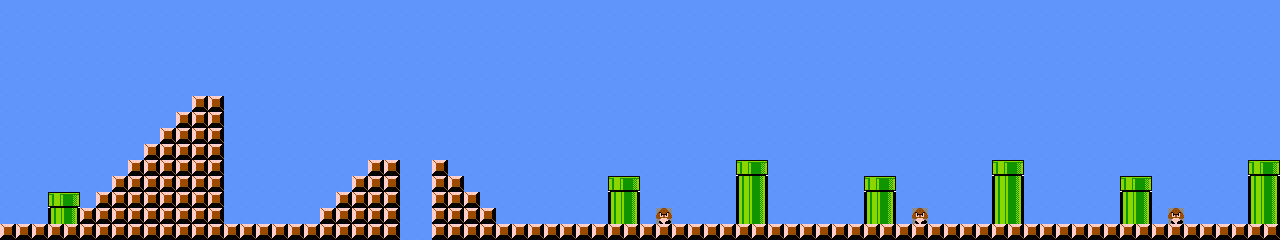} \\
\includegraphics[width=.3\textwidth]{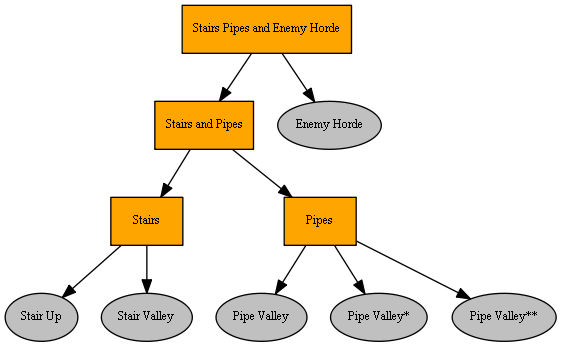} &
\includegraphics[width=.3\textwidth]{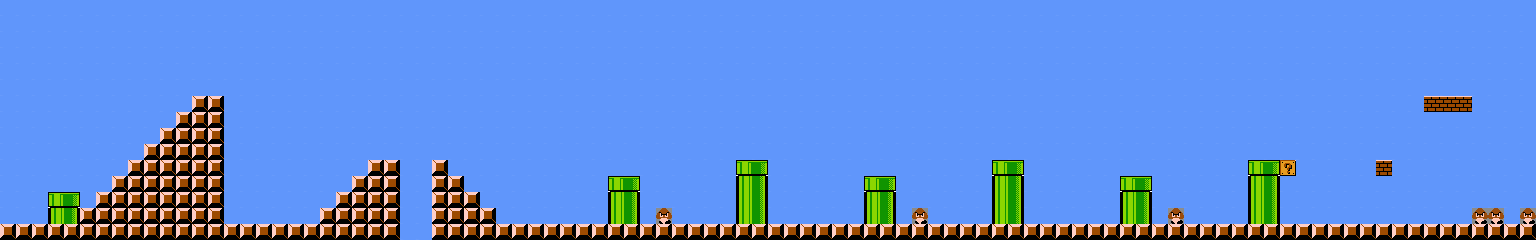} \\
\includegraphics[width=.4\textwidth]{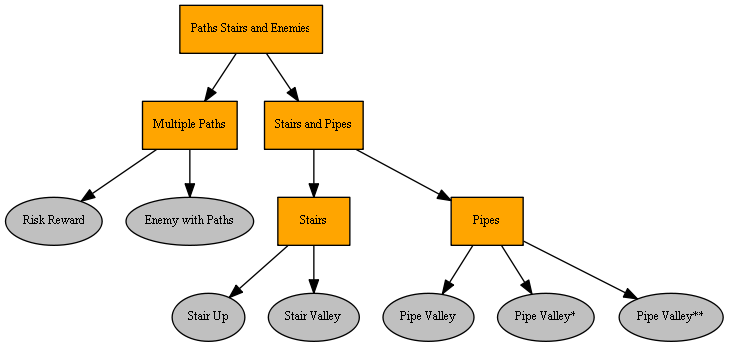} &
\includegraphics[width=.4\textwidth]{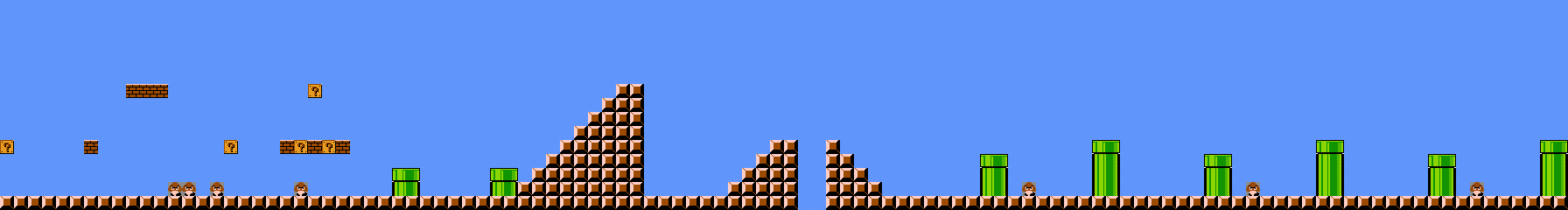} \\
\end{tabular}
\caption{\label{XFIGUREsmbworked} SMB worked example}
\end{figure}
}

\newcommand{\XFIGUREsmbwork}{
\begin{figure*}[h!]
\centering
\begin{subfigure}[t]{0.25\textwidth}
\centering
\includegraphics[width=0.25\textwidth]{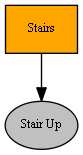}
\caption{Stair Up BT}
\end{subfigure}
~
\begin{subfigure}[t]{0.25\textwidth}
\centering
\includegraphics[width=0.25\textwidth]{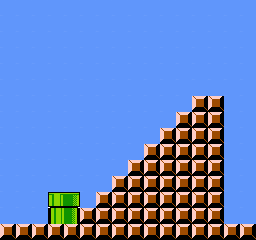}
\caption{Stair Up Level}
\end{subfigure}
\begin{subfigure}{0.25\textwidth}
\centering
\includegraphics[width=0.25\textwidth]{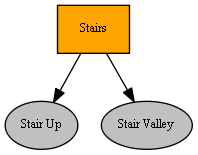}
\caption{StairUp StairValley BT}
\end{subfigure}
~
\begin{subfigure}{0.75\columnwidth}
\centering
\includegraphics[width=0.6\columnwidth]{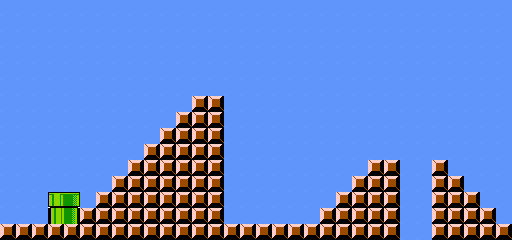}
\caption{StairUp StairValley Level}
\end{subfigure}
\begin{subfigure}{0.5\columnwidth}
\centering
\includegraphics[width=0.5\columnwidth]{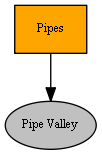}
\caption{Stair Up Stair Valley BT}
\end{subfigure}
~
\begin{subfigure}{0.45\columnwidth}
\centering
\includegraphics[width=0.45\columnwidth]{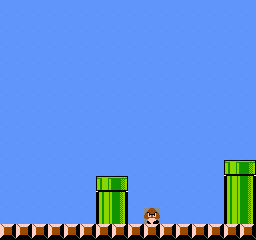}
\caption{Stair Up Stair Valley Level}
\end{subfigure}
\begin{subfigure}{0.75\columnwidth}
\centering
\includegraphics[width=0.5\columnwidth]{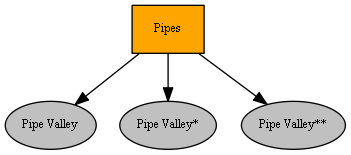}
\caption{Stair Up Stair Valley BT}
\end{subfigure}
~
\begin{subfigure}{0.75\columnwidth}
\centering
\includegraphics[width=0.5\columnwidth]{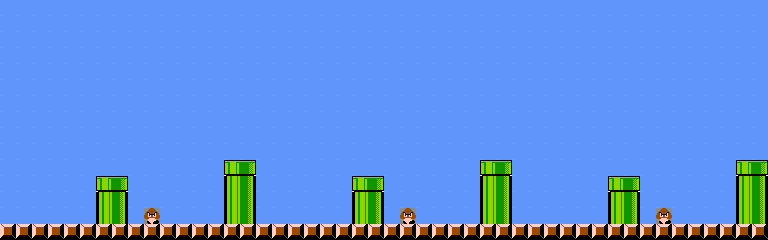}
\caption{Stair Up Stair Valley Level}
\end{subfigure}
\begin{subfigure}{0.45\columnwidth}
\centering
\includegraphics[width=0.45\columnwidth]{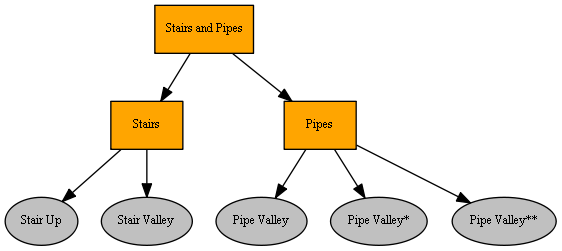}
\caption{Stair Up Stair Valley BT}
\end{subfigure}
~
\begin{subfigure}{0.45\columnwidth}
\centering
\includegraphics[width=0.45\columnwidth]{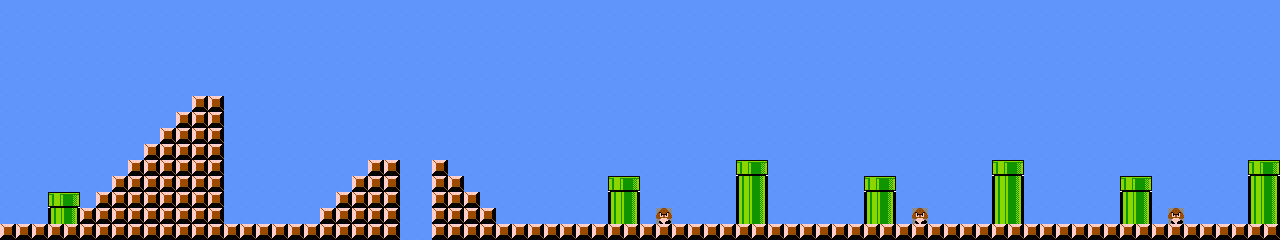}
\caption{Stair Up Stair Valley Level}
\end{subfigure}
\newline
\begin{subfigure}{0.45\columnwidth}
\centering
\includegraphics[width=0.45\columnwidth]{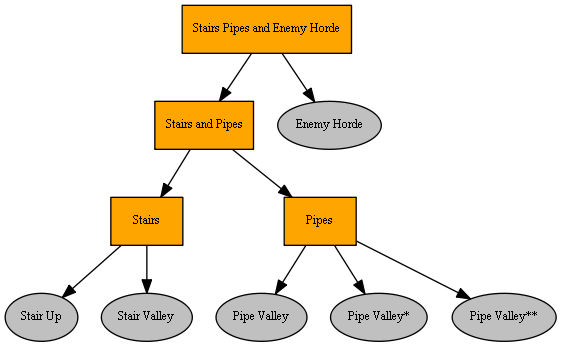}
\caption{Stair Up Stair Valley BT}
\end{subfigure}
~
\begin{subfigure}{0.45\columnwidth}
\centering
\includegraphics[width=0.45\columnwidth]{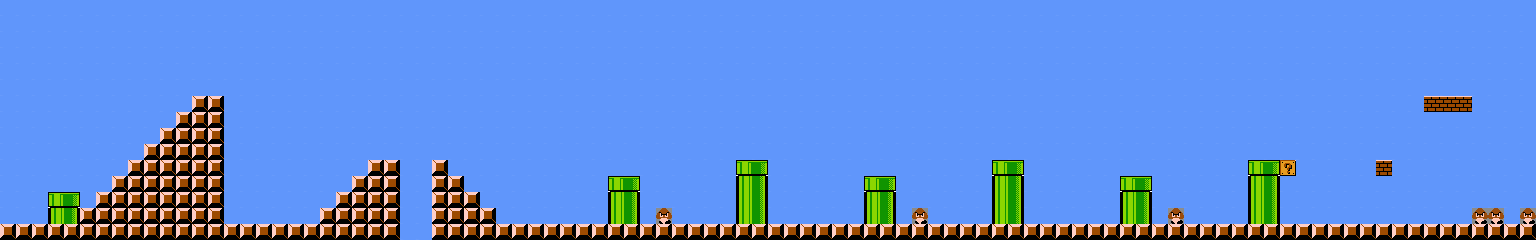}
\caption{Stair Up Stair Valley Level}
\end{subfigure}
\newline
\begin{subfigure}{0.45\columnwidth}
\centering
\includegraphics[width=0.45\columnwidth]{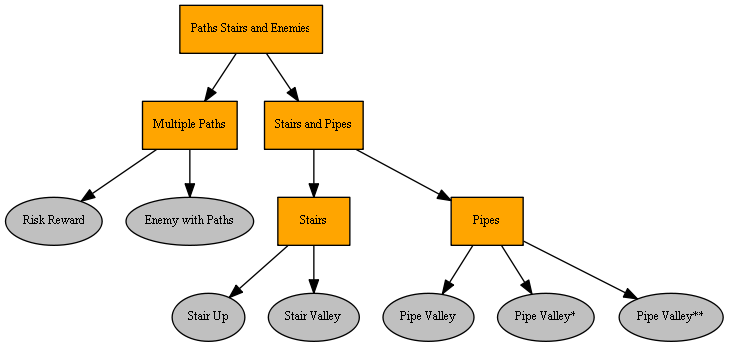}
\caption{Stair Up Stair Valley BT}
\end{subfigure}
~
\begin{subfigure}{0.45\columnwidth}
\centering
\includegraphics[width=0.45\columnwidth]{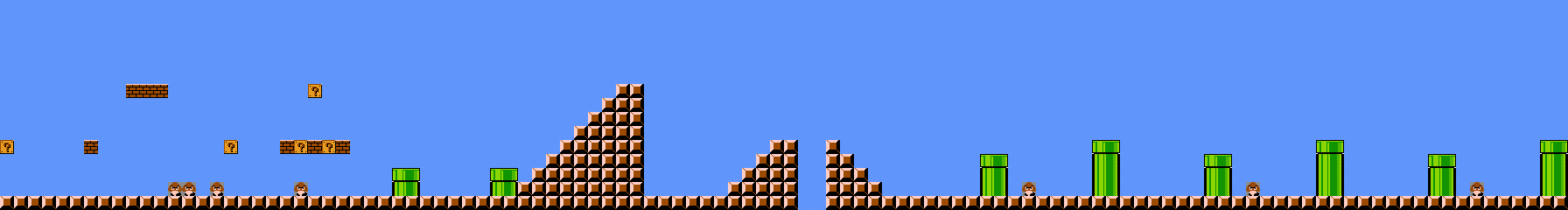}
\caption{Stair Up Stair Valley Level}
\end{subfigure}
\caption{\label{XFIGUREsmbwork} SMB worked example}
\end{figure*}
}

\newcommand{\XFIGUREmm}{
\begin{figure}[t]
\centering
\setlength\tabcolsep{1pt}
\small
\begin{tabular}{c"ccccc}
\includegraphics[width=0.15\columnwidth]{figure/mm_32_orig_00000} &
\includegraphics[width=0.15\columnwidth]{figure/mm_32_00001_1} &
\includegraphics[width=0.15\columnwidth]{figure/mm_32_00010_1} &
\includegraphics[width=0.15\columnwidth]{figure/mm_32_00100_1} &
\includegraphics[width=0.15\columnwidth]{figure/mm_32_01000_1} &
\includegraphics[width=0.15\columnwidth]{figure/mm_32_10000_1}
\\
\hline
\raisebox{2em}{\small{Random}} &
\includegraphics[width=0.15\columnwidth]{figure/mmz_32_00001} &
\includegraphics[width=0.15\columnwidth]{figure/mmz_32_00010} &
\includegraphics[width=0.15\columnwidth]{figure/mmz_32_00100} &
\includegraphics[width=0.15\columnwidth]{figure/mmz_32_01000} &
\includegraphics[width=0.15\columnwidth]{figure/mmz_32_10000}
\\
 & \clabel{00001} & \clabel{00010} & \clabel{00100} & \clabel{01000} & \clabel{10000} \\
\end{tabular}
\caption{\label{XFIGUREmm} MM segments generated by conditioning the original segment on the left (top) and a random vector (bottom) using the corresponding labels, as explained in Figure \ref{XFIGURElabels}.}
\end{figure}
}

\newcommand{\XFIGUREki}{
\begin{figure}[t]
\centering
\setlength\tabcolsep{1pt}
\small
\setlength\tabcolsep{1pt}
\begin{tabular}{c"cccc}
\includegraphics[width=0.16\columnwidth]{figure/ki_32_orig_0101} &
\includegraphics[width=0.16\columnwidth]{figure/ki_32_0001_2} &
\includegraphics[width=0.16\columnwidth]{figure/ki_32_0010_2} &
\includegraphics[width=0.16\columnwidth]{figure/ki_32_0100_2} &
\includegraphics[width=0.16\columnwidth]{figure/ki_32_1000_2}
\\
\hline
\raisebox{2em}{\small{Random}} &
\includegraphics[width=0.16\columnwidth]{figure/kiz_32_0001} &
\includegraphics[width=0.16\columnwidth]{figure/kiz_32_0010} &
\includegraphics[width=0.16\columnwidth]{figure/kiz_32_0100} &
\includegraphics[width=0.16\columnwidth]{figure/kiz_32_1000}
\\
 & \clabel{0001} & \clabel{0010} & \clabel{0100} & \clabel{1000} \\

\end{tabular}
\caption{\label{XFIGUREki} KI segments generated by conditioning the original segment on the left (top) and a random vector (bottom) using the corresponding labels as explained in Figure \ref{XFIGURElabels}.}
\end{figure}
}

\newcommand{\XFIGUREall}{
\begin{figure}[t]
\centering
\setlength\tabcolsep{1pt}
\small
\begin{tabular}{c"ccccc}
\includegraphics[width=0.15\columnwidth]{figure/smb_32_orig_11010} &
\includegraphics[width=0.15\columnwidth]{figure/smb_32_00001_1} &
\includegraphics[width=0.15\columnwidth]{figure/smb_32_00010_1} &
\includegraphics[width=0.15\columnwidth]{figure/smb_32_00100_1} &
\includegraphics[width=0.15\columnwidth]{figure/smb_32_01000_1} &
\includegraphics[width=0.15\columnwidth]{figure/smb_32_10000_1}
\\
\hline
\raisebox{2em}{\small{Random}} &
\includegraphics[width=0.15\columnwidth]{figure/smbz_32_00001} &
\includegraphics[width=0.15\columnwidth]{figure/smbz_32_00010} &
\includegraphics[width=0.15\columnwidth]{figure/smbz_32_00100} &
\includegraphics[width=0.15\columnwidth]{figure/smbz_32_01000} &
\includegraphics[width=0.15\columnwidth]{figure/smbz_32_10000}
\\
 & \clabel{000001} & \clabel{00010} & \clabel{00100} & \clabel{01000} & \clabel{10000} \\
\hline
& & & \textbf{(a) SMB} & & \\[3pt]
\hline
\hline
\includegraphics[width=0.16\columnwidth]{figure/ki_32_orig_0101} &
\includegraphics[width=0.16\columnwidth]{figure/ki_32_0001_2} &
\includegraphics[width=0.16\columnwidth]{figure/ki_32_0010_2} &
\includegraphics[width=0.16\columnwidth]{figure/ki_32_0100_2} &
\includegraphics[width=0.16\columnwidth]{figure/ki_32_1000_2} &
\\
\hline
\raisebox{2em}{\small{Random}} &
\includegraphics[width=0.16\columnwidth]{figure/kiz_32_0001} &
\includegraphics[width=0.16\columnwidth]{figure/kiz_32_0010} &
\includegraphics[width=0.16\columnwidth]{figure/kiz_32_0100} &
\includegraphics[width=0.16\columnwidth]{figure/kiz_32_1000}
& \\
 & \clabel{0001} & \clabel{0010} & \clabel{0100} & \clabel{1000} & \\
\hline
& & & \textbf{(b) KI} & & \\[3pt]
\hline
\hline
\includegraphics[width=0.15\columnwidth]{figure/mm_32_orig_00000} &
\includegraphics[width=0.15\columnwidth]{figure/mm_32_00001_1} &
\includegraphics[width=0.15\columnwidth]{figure/mm_32_00010_1} &
\includegraphics[width=0.15\columnwidth]{figure/mm_32_00100_1} &
\includegraphics[width=0.15\columnwidth]{figure/mm_32_01000_1} &
\includegraphics[width=0.15\columnwidth]{figure/mm_32_10000_1}
\\
\hline
\raisebox{2em}{\small{Random}} &
\includegraphics[width=0.15\columnwidth]{figure/mmz_32_00001} &
\includegraphics[width=0.15\columnwidth]{figure/mmz_32_00010} &
\includegraphics[width=0.15\columnwidth]{figure/mmz_32_00100} &
\includegraphics[width=0.15\columnwidth]{figure/mmz_32_01000} &
\includegraphics[width=0.15\columnwidth]{figure/mmz_32_10000}
\\
 & \clabel{00001} & \clabel{00010} & \clabel{00100} & \clabel{01000} & \clabel{10000} \\
 \hline
 & & & \textbf{(c) MM} & & \\[3pt]
 \hline
\end{tabular}
\caption{\label{XFIGUREall} \hspace{-0.1cm} SMB, KI and MM segments generated by conditioning the original segment on the left (top) and a random vector (bottom) using the corresponding labels, as explained in Figure \ref{XFIGURElabels}. Changing the label changes the content generated using the same the latent vector.}
\end{figure}
}

\newcommand{\XFIGURElabels}{
\begin{figure}[t]
\centering
\begin{tabular}{ccc}
\includegraphics[width=0.16\columnwidth]{figure/eg_smb_10011} &
\includegraphics[width=0.16\columnwidth]{figure/eg_ki_1101} &
\includegraphics[width=0.16\columnwidth]{figure/eg_mm_10101} \\
SMB - \clabel{10011} & KI - \clabel{1101} & MM - \clabel{10101}
\end{tabular}
\caption{\label{XFIGURElabels} Example original segments with corresponding element labels. \textit{Super Mario Bros} (SMB)-\clabel{\textit{Enemy, Pipe, Coin, Breakable, ?-Mark}}, \textit{Kid Icarus} (KI)-\clabel{\textit{Hazard, Door, Moving Platform, Fixed Platform}}, \textit{Mega Man} (MM)-\clabel{\textit{Hazards, Door, Ladder, Platform, Collectable}}. 0/1 in labels indicate absence/presence of corresponding elements in the segment.}
\end{figure}
}

\newcommand{\XFIGUREed}{
\begin{figure}[t!]
\centering
\includegraphics[width=0.8\columnwidth]{figure/cvae_ed_new}
\caption{\label{XFIGUREed} E-distances between original training distributions and distributions of 1000 levels generated using each of the blend conditioning labels.}
\end{figure}
}

\newcommand{\XFIGUREsmbplots}{
\begin{figure}[t]
\centering
\begin{tabular}{c}
\includegraphics[width=0.65\columnwidth]{figure/smb_both} \\
\includegraphics[width=0.65\columnwidth]{figure/smb_freq_new} \\
\end{tabular}
\caption{\label{XFIGUREsmbplots} Results of game element conditioning for both generated and training levels of SMB (top) with frequencies of each label in the training data (bottom). X-axis values are the integer encodings of the equivalent binary label. Results shown for 16 most frequent labels in the training levels.}
\end{figure}
}

\newcommand{\XFIGUREkiplots}{
\begin{figure}[t]
\centering
\begin{tabular}{c}
\includegraphics[width=0.65\columnwidth]{figure/ki_both} \\
\includegraphics[width=0.65\columnwidth]{figure/ki_freq_new} \\
\end{tabular}
\caption{\label{XFIGUREkiplots}Results of game element conditioning for both generated and training levels of KI (top) with frequencies of each label in the training data (bottom). X-axis values are the integer encodings of the equivalent binary label.}
\end{figure}
}

\newcommand{\XFIGUREmmplots}{
\begin{figure}[t]
\centering
\begin{tabular}{c}
\includegraphics[width=0.65\columnwidth]{figure/mm_both} \\
\includegraphics[width=0.65\columnwidth]{figure/mm_freq_new} \\
\end{tabular}
\caption{\label{XFIGUREmmplots} Results of game element conditioning for both generated and training levels of MM (top) with frequencies of each label in the training data (bottom). X-axis values are the integer encodings of the equivalent binary label. Results shown for 16 most frequent labels in the training levels.}
\end{figure}
}

\newcommand{\XFIGUREsmbpats}{
\begin{figure*}[h!]
\small
\vspace{-1.2cm}
\centering
\setlength\tabcolsep{1pt}
\begin{tabular}{c"cccccccc}
\includegraphics[width=0.16\columnwidth]{figure/pats_32_orig_0000101011} &
\includegraphics[width=0.16\columnwidth]{figure/pats_32_0000000010} &
\includegraphics[width=0.16\columnwidth]{figure/pats_32_0000001000}
&
\includegraphics[width=0.16\columnwidth]{figure/pats_32_0010101000} &
\includegraphics[width=0.16\columnwidth]{figure/pats_32_0100000000} &
\includegraphics[width=0.16\columnwidth]{figure/pats_32_0100001000} &
\includegraphics[width=0.16\columnwidth]{figure/pats_32_1000000000} &
\includegraphics[width=0.16\columnwidth]{figure/pats_32_1000001000} &
\includegraphics[width=0.16\columnwidth]{figure/pats_32_1100000000}
\\
\includegraphics[width=0.16\columnwidth]{figure/pats_32_orig_0111000000} &
\includegraphics[width=0.16\columnwidth]{figure/pats_32_0000000010_2} &
\includegraphics[width=0.16\columnwidth]{figure/pats_32_0000001000_2} &
\includegraphics[width=0.16\columnwidth]{figure/pats_32_0010101000_2} &
\includegraphics[width=0.16\columnwidth]{figure/pats_32_0100000000_2} &
\includegraphics[width=0.16\columnwidth]{figure/pats_32_0100001000_2} &
\includegraphics[width=0.16\columnwidth]{figure/pats_32_1000000000_2} &
\includegraphics[width=0.16\columnwidth]{figure/pats_32_1000001000_2} &
\includegraphics[width=0.16\columnwidth]{figure/pats_32_1100000000_2}
\\
\includegraphics[width=0.16\columnwidth]{figure/pats_32_orig_0000000000_4} &
\includegraphics[width=0.16\columnwidth]{figure/pats_32_0000000010_4} &
\includegraphics[width=0.16\columnwidth]{figure/pats_32_0000001000_4} &
\includegraphics[width=0.16\columnwidth]{figure/pats_32_0010101000_4} &
\includegraphics[width=0.16\columnwidth]{figure/pats_32_0100000000_4} &
\includegraphics[width=0.16\columnwidth]{figure/pats_32_0100001000_4} &
\includegraphics[width=0.16\columnwidth]{figure/pats_32_1000000000_4} &
\includegraphics[width=0.16\columnwidth]{figure/pats_32_1000001000_4} &
\includegraphics[width=0.16\columnwidth]{figure/pats_32_1100000000_4}
\\
\includegraphics[width=0.16\columnwidth]{figure/pats_32_orig_0000000000_5} &
\includegraphics[width=0.16\columnwidth]{figure/pats_32_0000000010_5} &
\includegraphics[width=0.16\columnwidth]{figure/pats_32_0000001000_5} &
\includegraphics[width=0.16\columnwidth]{figure/pats_32_0010101000_5} &
\includegraphics[width=0.16\columnwidth]{figure/pats_32_0100000000_5} &
\includegraphics[width=0.16\columnwidth]{figure/pats_32_0100001000_5} &
\includegraphics[width=0.16\columnwidth]{figure/pats_32_1000000000_5} &
\includegraphics[width=0.16\columnwidth]{figure/pats_32_1000001000_5} &
\includegraphics[width=0.16\columnwidth]{figure/pats_32_1100000000_5}
\\
Original & \clabel{\textit{SU}} & \clabel{\textit{MP}} & \clabel{\textit{PV-NV-MP}} & \clabel{\textit{G}} & \clabel{\textit{G-MP}} & \clabel{\textit{EH}} & \clabel{\textit{EH-MP}} & \clabel{\textit{EH-G}}
\end{tabular}
\caption{\label{XFIGUREsmbpats} Example segments generated by conditioning on SMB design patterns. The first segment in each row is from the game. Every other segment in its row is generated using the same vector as the original but conditioned using the label for that column. Results shown were generated using the 32-dimensional model. Labels indicate design patterns as defined previously.}
\end{figure*}
}

\newcommand{\XTABLEelems}{
\begin{table}[t]
\centering
\scriptsize
\setlength{\tabcolsep}{5pt}
\begin{tabular}{|c|c|c|c|c|c|c|}
\hline
\multicolumn{1}{|c|}{\multirow{2}{*}{}} & \multicolumn{2}{c|}{32-Dim} & \multicolumn{2}{c|}{64-Dim} & \multicolumn{2}{c|}{128-Dim} \\ \cline{2-7} 
\multicolumn{1}{|c|}{} & Exact & None & Exact & None & Exact & None \\ \hline
SMB-Rand & \textbf{33.5} & 17.6 & 32.7 & \textbf{17.1} & 27.1 & 19 \\ \hline
KI-Rand & \textbf{50.7} & \textbf{10} & 42 & 10.5 & 41.4 & 9.8 \\ \hline
MM-Rand & \textbf{17.4} & 43.2 & 15.2 & \textbf{42.5} & 14.5 & 43.7 \\ \hline
\hline
SMB-Train & \textbf{35.1} & \textbf{16.4} & 33.6 & \textbf{16.4} & 28.3 & 17.5 \\ \hline
KI-Train & \textbf{49.4} & \textbf{8.3} & 40.3 & 9.3 & 39.5 & 8.5 \\ \hline
MM-Train & \textbf{17.8} & 42.7 & 15.3 & \textbf{42.1} & 14.8 & 42.6 \\ \hline
\end{tabular}
\caption{\label{XTABLEelems} Results of conditioning randomly sampled (`Rand') and training (`Train') segments using element labels. Highest Exact values and lowest None values per game are highlighted in bold.}
\end{table}
}

\newcommand{\XTABLEelemsrand}{\begin{table}[t]
\centering
\small
\setlength{\tabcolsep}{5pt}
\begin{tabular}{|c|c|c|c|c|c|c|}
\hline
& \multicolumn{2}{c}{32-dim} & \multicolumn{2}{|c}{64-dim } & \multicolumn{2}{|c|}{128-dim} \\
\hline
 & Exact & None & Exact & None & Exact & None \\
\hline
SMB & \textbf{33.49} & 17.6 & 32.73 & \textit{17.09} & 27.05 & 18.98\\
\hline
KI & \textbf{50.74} & 9.89 & 42 & 10.46 & 41.44 & \textit{9.76}\\
\hline
MM & \textbf{17.38} & 43.24 & 15.24 & \textit{42.53} & 14.54 & 43.67\\
\hline
\end{tabular}
\caption{\label{XTABLEelemsrand} Results conditioning random vectors using element labels. Highest Exact values and lowest None values per game are highlighted in bold and italics respectively.}
\end{table}
}

\newcommand{\XTABLEelemstrain}{\begin{table}[t!]
\centering
\small
\setlength{\tabcolsep}{5pt}
\begin{tabular}{|c|c|c|c|c|c|c|}
\hline
& \multicolumn{2}{c}{32-dim} & \multicolumn{2}{|c}{64-dim } & \multicolumn{2}{|c|}{128-dim} \\
\hline
 & Exact & None & Exact & None & Exact & None \\
\hline
SMB & \textbf{35.05} & 16.43 & 33.59 & \textit{16.42} & 28.34 & 17.52\\
\hline
KI & \textbf{49.4} & \textit{8.29} & 40.32 & 9.25 & 39.47 & 8.52\\
\hline
MM & \textbf{17.75} & 42.68 & 15.27 & \textit{42.13} & 14.75 & 42.62\\
\hline
\end{tabular}
\caption{\label{XTABLEelemstrain}  Results of conditioning training levels using element labels. Highest Exact values and lowest None values per game are highlighted in bold and italics respectively.}
\end{table}
}

\newcommand{\XTABLEblend}{\begin{table}[t!]
\scriptsize
\centering
\setlength{\tabcolsep}{4pt}
\begin{tabular}{|c||c|c|c||c|c|c||c|c|c|}
\hline
& \multicolumn{3}{c||}{32-dim CVAE} & \multicolumn{3}{c||}{64-dim CVAE} & \multicolumn{3}{c|}{128-dim CVAE} \\
\hline
 Label & SMB & KI & MM & SMB & KI & MM & SMB & KI & MM \\
\hline
\clabel{000} &  38.7 & 18.1 & \textbf{43.2} & 31 & 20.3 & \textbf{48.7} & \textbf{41.5} & 18.2 & 20.3\\
\hline
\clabel{001} &  3.8 & 2.4 & \textbf{93.8} & 2.7 & 3.7 & \textbf{93.6} & 3.5 & 2.9 & \textbf{93.6}\\
\hline
\clabel{010} &  0.7 & \textbf{95.5} & 3.8 & 1.5 & \textbf{93.6} & 4.9 & 0.7 & \textbf{94.5} & 4.8\\
\hline
\clabel{011} &  6.8 & 22.9 & \textbf{70.3} & 7.8 & 27.5 & \textbf{64.7} & 10 & 24 & \textbf{66}\\
\hline
\clabel{100} &  \textbf{97.6} & 1.4 & 1 & \textbf{98.8} & 1.1 & 0.1 & \textbf{98.9} & 0.7 & 0.4\\
\hline
\clabel{101} & \textbf{71.9} & 2.9 & 25.2 & 20.7 & 5.2 & \textbf{74.1} & 38.1 & 2.6 & \textbf{59.3}\\
\hline
\clabel{110} &  \textbf{86.5} & 11.8 & 1.7 & \textbf{59} & 34.5 & 6.5 & \textbf{57.4} & 33.5 & 9.1\\
\hline
\clabel{111} & \textbf{56.7} & 10.3 & 33 & 32.1 & 16.8 & \textbf{51.1} & \textbf{45} & 11.1 & 43.9\\
\hline
\end{tabular}
\caption{\label{XTABLEblend} For each label, percentage of blended segments generated using that label, that was classified as the different games. Highest percentage classification for each label-dimensionality pair highlighted in bold.}
\end{table}
}

%% file: body.tex
\section{Introduction}
Behavior trees (BTs) \cite{isla2005handling, champandard2007behavior} are a commonly used technique and framework for modeling agent behaviors in games and have seen widespread use in a large number of commercial games for defining NPC and enemy AI. The popularity of BTs stems from them enabling such behaviors to be implemented in a modular and reactive manner. That is, simpler actions and behaviors can be combined to define more complex agent behaviors and different BT branches corresponding to different behaviors can be selected for execution based on various runtime conditions, thus enabling agents to react dynamically during gameplay. In addition to game AI, in recent years, BTs have also been widely employed in robotics for defining behaviors of robot controllers \cite{colledanchise2017behavior}, thus demonstrating the general usefulness of BT-based methods.

Hence in this paper, with a view to bring these benefits to bear on generative methods, we propose the use of BTs for procedural content generation (PCG). More specifically, we repurpose BTs to model the behavior of game \textit{design} agents rather than game \textit{playing} agents. By replacing NPC actions (e.g. Cover and Shoot) with design actions (e.g. Generate Segment and Connect Rooms), we obtain BTs that are procedural level generators and are also modular, reactive and interpretable like their traditional BT counterparts. We refer to this framework as Procedural Content Generation using BTs (PCGBT) and BTs capable of doing so as PCG-BTs. We demonstrate this approach by developing PCG-BTs for generating \textit{Super Mario Bros.} and \textit{Mega Man} levels as well as dungeon layouts. We also show that BTs can be used to describe a generic level generator which can then be instantiated for different games such as \textit{Mega Man} and \textit{Metroid}. Further, we show that subtrees corresponding to different games could be combined to form a super-tree capable of generating levels that blend portions of different games. Finally, we conclude with an extensive discussion on future applications and implications of this PCGBT framework. Our work thus contributes, to our knowledge, the first application of BTs for explicitly defining game design agents, and thereby enabling procedural content generation.

\section{Background}
Most prior BT-related research has focused on using BTs for modeling game playing agents and controllers rather than PCG. Evolving BTs for controlling agents has seen much work with \citeauthoryearp{lim2010evolving} focusing on the real-time strategy game DEFCON and \citeauthoryearp{perez2011evolving} and \citeauthoryearp{nicolau2016evolutionary} evolving controllers for the Mario AI framework. Instead of evolution, \citeauthoryearp{robertson2015building} identified patterns in action sequences for producing BTs for \textit{StarCraft} while \citeauthoryearp{glavin2014adaptive} used a reinforcement learning-based approach for building BTs for bots in \textit{Unreal Tournament 2004}. In recent years, robotics has arguably seen more BT research than games. Comprehensive surveys of BTs in robotics have been given by \citeauthoryearp{iovino2020survey} and 
\citeauthoryearp{colledanchise2017behavior}. 

In using tree-like structures to model generators, our work overlaps with graph and grammar-based generative approaches. \citeauthoryearp{shaker2012evolving} used grammatical evolution to evolve graphs of Mario levels while \citeauthoryearp{dormans2010adventures} and \citeauthoryearp{karavolos2015mixed} both utilized mission and space graphs for dungeon generation. Relatedly, graph grammars have been utilized for generating levels for Mario \cite{hauck2020automatic}, puzzle games \cite{valls2017graph}, dungeon crawlers \cite{linden2013designing} and educational games \cite{jemmali2020grammar}. Rather than graphs, we use an explicit BT formulation for defining level generators. The recently introduced field of PCG via Reinforcement Learning (PCGRL) \cite{khalifa2020pcgrl,nam2019generation} also seeks to produce game design agents which perform generative actions. Instead of RL, we use BTs to model design agents.


\section{Method}

In this section we describe our approach. We first describe how BTs work and then detail how we adapt them for PCG.

\subsection{Behavior Tree (BT)}
A BT is a directed tree structure consisting of a root node where execution starts, a set of internal (or control flow) nodes which control the flow of execution and a set of leaf nodes which define the actions to be executed. The root controls execution by propagating signals called ticks at a pre-determined frequency. Each node receives ticks from its parent node and propagates it to one or more of its child nodes. A node executes only if it receives a tick from its parent. Each child, after finishing execution, returns one of 3 statuses to its parent---\textit{Running}, \textit{Success} or \textit{Failure}---which in turn determines the status of the parent. There are typically 4 types of control flow nodes:
\begin{itemize}
    \item \textit{Sequence} - execute all their children in order from left to right until one fails. They return \textit{Success} only if all of their children also succeed. 
    \item \textit{Selector} - execute all their children in order from left to right until one of them succeeds. They return \textit{Failure} only if all of their children fail.
    \item \textit{Parallel} - execute their children simultaneously and succeed if a predetermined number of children succeed. 
    \item \textit{Decorator} - modify their child behavior node.
\end{itemize}

Leaf nodes are typically of 2 types:
\begin{itemize}
    \item \textit{Action} - execute commands corresponding to the most low-level behaviour being modeled.
    \item \textit{Condition} - used to check conditions and return success or failure accordingly, corresponding to true and false.
\end{itemize}

More working and implementation details about BTs can be found in
\citeauthoryearp{champandard2019behavior} and \citeauthoryearp{colledanchise2017behavior}. Our PCGBT approach is enabled by having \textit{Action} nodes perform level design tasks rather than NPC/enemy behaviors. In this work, we only used \textit{Sequence} and \textit{Selector} control flow nodes since they were sufficient to demonstrate the application of PCG-BTs but \textit{Parallel} and \textit{Decorator} nodes could also be incorporated in the future. Further, all nodes in this work returned either \textit{Success} or \textit{Failure} though in the future these could return \textit{Running}, while waiting for playability checks to complete, for example. In general, all features of traditional BTs would apply to PCG-BTs but due to limited space and wanting to demonstrate PCG-BTs for multiple games, we focused on a subset of BT features. In all cases, unless indicated, \textit{Condition} nodes returned \textit{Success} or \textit{Failure} at random, i.e. if a randomly generated float was less than 0.5, we returned \textit{Success}, else \textit{Failure}. The result was then propagated back up and branches chosen accordingly, as per typical BT rules. In all figures in this work, \textit{Sequence}, \textit{Selector}, \textit{Action} and \textit{Condition} nodes are represented as orange rectangles, blue octagons, grey ellipses and grey diamonds respectively.

BTs may optionally also use a blackboard. In its simplest form, a blackboard is a key-value store that is globally accessible by all nodes in the tree and used to store information that may be useful when making decisions. We make extensive use of blackboards in our BT implementations.

\XFIGUREsmbworkedfix


\subsection{Level Generation using BTs}
Analogous to how traditional BTs work with a library of scripted actions which are then combined in different ways to define different behaviors, the PCG-BTs that we use work with a library of level segments, which are combined to generate different levels. That is, executing an action node in a PCG-BT places a level segment of a certain type as defined by that node with different action nodes corresponding to placing different types of segments. We also make use of global blackboard variables to keep track of the position in the level where the next segment should be placed. During execution, each action node updates this position after placing its node, with updates varying from game to game, e.g., in Mario, updates increment the x-coordinate by 1 whereas in Mega Man, this depends on if the next segment should be placed above, below or to the right of the current segment. The blackboard is particularly useful for PCGBT as it can store globally accessible information useful for aiding the generative process such as designer preferences for selecting specific node types, player data to determine which branches to execute, the state of the current game world, etc. 

To illustrate how we use BTs to model design agents, we go through a step-by-step worked example for generating a Mario level, as shown in Figure \ref{XFIGUREsmbworkedfix}. In this work, we use Mario design patterns as defined by \citeauthoryearp{dahlskog2012patterns} as the building blocks of levels with different action nodes corresponding to the placement of segments with different design patterns. We start with a very simple tree with 1 leaf node that places a \textit{Stair Up} segment. This is then combined with another leaf that generates a \textit{Stair Valley} to give us a tree that generates a section consisting of 2 stair segments. We define a separate tree that generates a section of 3 \textit{Pipe Valleys}. Next, we combine these 2 separate BTs to obtain a larger tree that can now generate a stair section followed by a section of pipe valleys. This larger BT could be further expanded in either direction as shown in the next examples where we add an enemy section to the right and then a section of multiple paths to the left. This example demonstrates how we can build up BTs for generating full levels by combining smaller BTs that generate level sections, thus yielding modular level generators, similar to how traditional BTs enable modeling agent behavior in a modular fashion. In the next section, we demonstrate applications of BTs for developing level generators for a number of different games.


\section{Applications}

In this section, we demonstrate applications of PCG-BTs for several games. In all cases, we used levels from the Video Game Level Corpus \cite{summerville2016vglc}. BTs were implemented using py\_trees\footnote{\url{https://github.com/splintered-reality/py_trees}}.

\XFIGUREsmboneone

\XFIGUREsmblevelbt

\subsection{Super Mario Bros. (SMB)}
For SMB, we created the library of level segments for the BTs by extracting non-overlapping 14x16 segments from VGLC levels. Segments were then padded with 1 row at the top for consistency in combining with segments from other games as described later. We manually categorized the segments based on the design patterns within them. Action nodes were defined to take one or more design patterns as parameters and sample a segment from the set of segments that contain at least one of the desired pattern(s). Different action nodes could then be combined to form subtrees capable of generating level sections with different patterns, which in turn could be combined to form larger trees capable of generating entire levels. To keep track of the level during generation, we maintained a globally accessible dictionary on the blackboard, mapping (x,y) coordinates to the segment generated for that location. At the end of execution, each action node incremented x by 1 and added the sampled segment to this dictionary. We show two example BTs with the first recreating and also generating variations of level 1-1. We model this level as a BT with 3 top-level sequence nodes---the first generates the initial section of pipes, the second generates the middle section of enemies and multiple paths and the third generates the final stair-filled section. The BT and 2 sample levels are shown in Figure \ref{XFIGUREsmboneone}. The first `verbatim' sample reproduces the original level by simply copying the corresponding segment rather than sampling, i.e. we hard-coded action nodes to pick the exact segment that appears in the corresponding position in the original level. In the second `sampled' example, the action node samples a segment with the same design pattern in the corresponding position in the original level. These examples show that BTs can be used to sufficiently model generators capable of producing existing levels as well as their variations. The second example BT and the resulting sampled levels generated using it are shown in Figure \ref{XFIGUREsmblevelbt}. This example demonstrates the use of \textit{Selector} nodes for picking between two branches of execution. The first decides whether to generate a section of paths and pipes or stairs and enemies while the second decides between a section with gaps or valleys. The decision to branch is made in the \textit{Do Path-Pipe?} and \textit{Do Gap?} condition leaf nodes. For these examples, these choices were made randomly but one could have selection probabilities be weighted by designer preferences to guide selection. Also, in a dynamic setting, player data could be used to make the choice. For example, in \textit{Do Gap?}, we could check if the player has already lost a number of lives and generate valleys instead of gaps to reduce difficulty. We note that these BTs generated the entire level in one tick from the root rather than a repeating loop of ticks like BTs employed at runtime.

\subsection{Mega Man (MM)}
For MM, we constructed a similar library of 15x16 segments from the VGLC but grouped segments based on the directions in which they were open rather than design patterns, since this information is crucial for combining segments such that the overall level is traversable, e.g. a segment that is open above needs the next segment to be open below. Each segment was tagged with U, D, L and/or R to indicate it being open above, below, to the left and/or to the right respectively. Similar to SMB action nodes, MM action nodes sampled segments based on the desired opening orientations. Also, since MM levels do not exclusively progress to the right like SMB, each action node also performed a check to see if there was an opening between the current segment being sampled and the previously sampled segment along the edge adjoining them. This ensured generating segments such that the resulting level has a traversable path through it. Similar to the SMB BT, we maintained a level dictionary on the blackboard in addition to a reference to the previous segment for performing the aforementioned check. After sampling its segment, each action node updated both the x and y coordinates based on the position of the next segment. For space, we only show one MM BT in Figure \ref{XFIGUREmmlevelbt}. This generates an initial horizontal section then decides between generating a horizontal or a vertical section followed by a middle horizontal section, a vertical section and a final horizontal section. Note also that each vertical section is generated by a \textit{Selector} that decides whether the section will progress upward or downward. Similar to SMB, these decisions were made at random but could be made based on designer preferences or at runtime based on player behavior. Also like SMB, the BT here used a single root tick to generate the entire level.

The SMB and MM BTs both produced levels in only a linear manner due to the original games lacking loops and progressing linearly, necessitating a linear generation approach. PCG-BTs can also generate levels that progress in multiple directions and have paths that loop around, as we'll see next.

\XFIGUREmmlevelbt

\subsection{Dungeons}
Having demonstrated BTs for platformer level generation using a segment-by-segment generative approach, we wanted to test BTs for generating levels for other genres. Thus, we implemented BTs for generating layouts for dungeon crawlers such as \textit{The Legend of Zelda}. We developed a BT to model a simple layout generation algorithm which begins by placing a starting room that is closed on all sides. Then in each iteration, a random closed side is chosen. If there is no room next to it, a room is placed there with all its sides closed. The two rooms are then connected by setting the adjoining edge to be open. This is repeated until the desired number of rooms has been generated. To visualize the generated dungeons, we extracted the 11x16 rooms from Zelda levels in the VGLC. Similar to MM, we tagged each room with the direction in which they had doors. For each room generated by the BT, we placed a sampled Zelda room with doors in the corresponding directions. The BT and sample dungeons are shown in Figure \ref{XFIGUREdungeonbtacross}. Unlike previously, the BT here generates one room per tick and executes a loop of ticks until all rooms have been generated. For this, we store a flag on the blackboard to indicate if the start room has been generated, initialized to false. During the first tick, the start \textit{Selector} executes, generates the starting room and disables the flag so that on subsequent ticks this check fails. Note that the SMB and MM BTs could also have been implemented using a loop of ticks and the dungeon could have been implemented using a single tick, demonstrating the different ways in which BTs could be used for level generation.


\XFIGUREdungeonbtacross

\XFIGUREgenericbtacross

\subsection{Generic and Blending}
In addition to modeling specific games, we envisioned crafting game-agnostic BTs which could be instantiated using levels from different games. This only requires that sections generated by the control flow nodes and segments produced by the action nodes are compatible across multiple games. We have seen two such BTs. While the SMB BT is not generic since it utilizes Mario-specific design patterns, the MM BT can generate levels for any platformer consisting of vertical and horizontal sections and segments with openings in 4 directions. Similarly, the dungeon layout BT could be used for any game with interconnected segments. \textit{Metroid}, being a platformer and having a sprawling interconnected game world, satisfies both criteria. Similar to MM, we used 15x16 segments from full Metroid levels from the VGLC. Segments were similarly labeled based on directionality with slight modifications where needed (e.g. no Metroid segment had just Down and Right openings without Up also being open, thus we mapped the DR action node to sample UDR in this case). An example Metroid level generated using the dungeon BT is shown in Figure \ref{XFIGUREdungeonbtacross}. A generic BT adapted from the prior MM BT is shown in Figure \ref{XFIGUREgenericbtacross} along with an example MM level and Metroid level, both generated using it. Finally, given that subtrees corresponding to different level sections could be combined to form larger trees for generating whole levels, one could combine a tree for generating levels of one game with a tree for generating levels of another to produce a more complex tree that can generate levels from different games taken together i.e., perform game blending \cite{gow2015towards}. Such a blending BT is shown in Figure \ref{XFIGUREblendbt}. It consists of an initial \textit{Sequence} for generating an SMB section, followed by a \textit{Selector} for generating either a horizontal or a vertical MM section, followed by a \textit{Sequence} for generating a horizontal and then a vertical section of Metroid. Playability is maintained similar to MM action nodes i.e. by checking adjoining edges of successive segments and resampling if no path exists.

\XFIGUREblendbt

\section{Discussion and Future Directions}

Our results suggest that BTs can be repurposed for modeling design agents and yield modular and interpretable generators---modular in that subtrees generating level sections can be recombined with other subtrees to produce a variety of BTs and interpretable in that BTs indicate why and how the generators produce the levels they do. Note that we are not proposing a specific algorithm or BT implementation but simply the use of BTs for modeling procedural level generators rather than NPCs, and demonstrating that it is feasible to do so via our examples. The implementations of the underlying condition and action nodes are agnostic to the PCGBT framework, e.g. all selection decisions in our examples are made at random. As mentioned before, these decisions could also be made based on designer preferences or player behavior at runtime. Decoupling the PCGBT framework from underlying implementations enables the approach to generalize to multiple design styles/preferences without enforcing any one. The primary utility of PCG-BTs would be to let designers combine sub-levels or handmade content into whole levels in a modular, explainable manner. Overall, we believe this initial exploratory foray into defining and using PCG-BTs holds promise for several interesting future directions.

\subsubsection{Dynamic Level Generation} Similar to traditional BTs, PCG-BTs could be reactive and produce dynamic generators where different level sections are generated by choosing different branches based on runtime conditions. Moreover, by having conditions tied to player behavior, we could generate different levels and gameplay experiences specifically tailored towards different player types and difficulty levels. 

\subsubsection{Combining with traditional BTs} 
Modeling both design and gameplay agents via the same formalism opens up many generative possibilities. We could develop hybrid BTs with subtrees for both agent AI and level generation. Execution of a level generation branch could be controlled based on if a game playing branch succeeds or fails. Prior work \cite{cooper2020pathfinding} looked at using pathfinding agents for level repair. Such agents in BT form could be combined with PCG-BTs. There are also several directions using evolutionary algorithms. Could we meaningfully crossover and mutate such hybrid trees? Would we find commonalities between BT agents able to play certain PCG-BTs? Could we co-evolve game playing BTs and PCG-BTs in an open-ended manner to discover new types of agents and games?

\subsubsection{RL and Evolution} We used hand-crafted PCG-BTs but in the future they could be generated via learning and evolution. Prior work has used RL to learn BTs \cite{dey2013ql,banerjee2018autonomous} and could be leveraged for learning agents capable of designing games, akin to PCGRL. Evolution has also been applied for evolving BTs modeling desired behaviors \cite{jones2018evolving,neupane2019learning}, so we could evolve PCG-BTs modeling desired levels. Both RL and evolution could be used by designers for inferring BT structures from a set of exemplar levels.


\subsubsection{General Game Design} We showed that BTs can model game-agnostic level generators which can be instantiated to produce levels for different games and genres using the same tree. Thus PCG-BTs are capable of general level generation \cite{togelius2016general}, i.e. generating levels for a number of different games. This could in turn enable dynamic game generation analogous to dynamic level generation, producing different games for different players. Further, we could incorporate the blend BTs, and dynamically switch different games in and out during gameplay. In the future, it would be interesting to develop such generalized PCG-BTs. For e.g., could we cluster similar PCG-BTs into \textit{forests} of BTs capable of generating similar types of games?

\subsubsection{Designers and Practitioners} PCG techniques, particularly those involving evolution, ML \cite{summerville2017procedural} and RL have not been widely adopted in commercial games. \citeauthoryearp{jacob2020s} point out that practitioners seek methods that afford authorial control, are readable and easy to interpret and do not require research expertise. Since BTs are a very well-known AI technique among designers, PCG-BTs could be a more accessible generative approach to those outside of research circles.

\section{Conclusion}
We introduced PCGBT, i.e., the use of behavior trees for procedural content generation, and demonstrated its application in several games and use-cases. While we did not perform playability evaluations on generated levels, we note that the building blocks of our levels are taken directly from the original games and thus segments themselves are playable. For dungeons, the layout algorithm guarantees that the dungeon is traversable from one room to another. Of course, if our action nodes explicitly generated the segments from scratch, playability evaluations of those would be necessitated. Since we focused on investigating the feasibility of this approach, we leave such considerations of overall playability of levels for future work. Moreover, in addition to the directions discussed previously, in the future, we want to develop a GUI/interactive application to enable designers to create and combine their own PCG-BTs as well as conduct a user study to examine PCGBT-generated levels in a dynamic, reactive context.